\documentclass[final,12pt,nosubfloats]{clear2023} %

\usepackage[utf8]{inputenc}
\usepackage[T1]{fontenc}
\usepackage{xcolor}
\usepackage{microtype}
\usepackage[parfill]{parskip}
\usepackage{hyperref}
\usepackage{amsfonts}%
\usepackage{booktabs}
\usepackage{nicefrac}
\usepackage{bm}
\usepackage{xpatch}
\usepackage{tikz}
\usepackage{algorithm}
\usepackage[noend]{algpseudocode}
\usepackage[capitalize,noabbrev]{cleveref}
\usepackage{subcaption}
\usepackage{wrapfig}
\usepackage{comment}
\usepackage[inline,shortlabels]{enumitem}

\title[Stochastic Causal Programming]{Stochastic Causal Programming for Bounding Treatment Effects}
\usepackage{times}

\graphicspath{{figs/}}

\makeatletter
\xpatchcmd{\algorithmic}{\itemsep\z@}{\itemsep=0.5ex plus1pt}{}{}
\makeatother

\usetikzlibrary{cd,arrows,calc,shapes,positioning}
\tikzstyle{obs} = [circle,fill=white,draw=black,inner sep=0pt,minimum size=18pt,font=\fontsize{10}{10}\selectfont,node distance=1,thick]
\tikzstyle{latent} = [obs,dotted]

\newcommand{\edge}[3][]{ %
  \foreach \x in {#2} { %
    \foreach \y in {#3} { %
      \path (\x) edge [->, >={Latex[round]}, #1,thick] (\y) ;%
    } ;
  } ;
}

\definecolor{lightgrey}{rgb}{0.925, 0.925, 0.925}
\newcommand{\commentbox}[1]{\colorbox{lightgrey}{#1}}
\newcommand{\trueline}{\raisebox{2pt}{\tikz{\draw[black,solid,ultra thick](0, 0) -- (5mm, 0);}}}
\newcommand{\regressline}{\raisebox{2pt}{\tikz{\draw[black,dashed,ultra thick](0, 0) -- (5mm, 0);}}}
\newcommand{\boundline}{\raisebox{2pt}{\tikz{\draw[dotted,ultra thick](0, 0) -- (6mm, 0);}}}

\newcommand{\xhdr}[1]{\textbf{#1.}\;}

\newcommand{\bR}{\ensuremath \mathbb{R}}

\newcommand{\cF}{\ensuremath \mathcal{F}}
\newcommand{\cL}{\ensuremath \mathcal{L}}
\newcommand{\cS}{\ensuremath \mathcal{S}}
\newcommand{\cX}{\ensuremath \mathcal{X}}
\newcommand{\cY}{\ensuremath \mathcal{Y}}
\newcommand{\cZ}{\ensuremath \mathcal{Z}}
\newcommand{\cN}{\ensuremath \mathcal{N}}
\newcommand{\cD}{\ensuremath \mathcal{D}}

\DeclareMathOperator{\E}{\mathbb{E}}
\DeclareMathOperator{\dist}{\mathrm{dist}}
\DeclareMathOperator{\indep}{\perp\!\!\!\perp}

\newcommand{\B}[1]{\bm{#1}}
\newcommand{\given}{\,|\,}
\newcommand{\lhs}{B}
\newcommand{\rhs}{A}
\newcommand{\obj}{\texttt{[obj]}}
\newcommand{\cdata}{\texttt{[c-data]}}
\newcommand{\cstruct}{\texttt{[c-struct]}}

\clearauthor{%
 \Name{Kirtan Padh} \Email{kirtan.padh@helmholtz-muenchen.de}\\
 \addr Helmholtz AI, Helmholtz Munich \& Technical University Munich
 \AND
 \Name{Jakob Zeitler}\\
 \addr University College London
 \AND
 \Name{David Watson}\\
 \addr King's College London
 \AND
 \Name{Matt Kusner}\\
 \addr University College London
 \AND
 \Name{Ricardo Silva}\\
 \addr University College London
 \AND
 \Name{Niki Kilbertus}\\
 \addr Helmholtz AI, Helmholtz Munich \& Technical University Munich
}

\begin{document}

\maketitle

\begin{abstract}%
Causal effect estimation is important for many tasks in the natural and social sciences. We design algorithms for the continuous \emph{partial identification} problem: bounding the effects of multivariate, continuous treatments when unmeasured confounding makes identification impossible. Specifically, we cast causal effects as objective functions within a constrained optimization problem, and minimize/maximize these functions to obtain bounds. We combine flexible learning algorithms with Monte Carlo methods to implement a family of solutions under the name of \emph{stochastic causal programming}. In particular, we show how the generic framework can be efficiently formulated in settings where auxiliary variables are clustered into \emph{pre-treatment} and \emph{post-treatment sets}, where no fine-grained causal graph can be easily specified. In these settings, we can avoid the need for fully specifying the distribution family of hidden common causes. Monte Carlo computation is also much simplified, leading to algorithms which are more computationally stable against alternatives.
\end{abstract}


\section{Introduction}
\label{sec:intro}

Estimating causal effects is a key goal of scientific inquiry \citep{pearl2009causality, Imbens2015, hernan2020}. When unobservable confounders between some treatment $X$ and some outcome $Y$ are present, observed \emph{auxiliary} variables can be exploited in a variety of identification strategies, such as (a) instruments in instrumental variable (IV) models \citep[Ch. 16]{hernan2020}; (b) mediators in the front-door criterion \citep[Ch. 3]{pearl2009causality}; or (c) noisy measurements of the confounders, known as proxies \citep{tchetgen2020}. 

Even then, identification requires further assumptions. Assumptions are needed not just about the graphical structure, but also about the functional form of causal associations, e.g., monotonicity \citep{angrist1995}, additivity \citep{hartford2017deep, singh2019kernel, muandet2019dual, bennet2019} or conditions such as completeness \citep{tchetgen2020}. Some of these assumptions are not immediately intuitive, as they refer to undetermined hidden variables lacking interpretation. %
When such assumptions fail, effects may still be \emph{partially identifiable}. That is, we can report a non-vacuous \emph{set} containing all effects implied by the causal models consistent with the data and assumptions \citep{Manski1990}. Effect sets are typically described as intervals, which we can define in terms of a lower bound and an upper bound on the causal quantity of interest. In this scenario, using Pearl's \emph{do}-operator notation \citep{pearl2009causality}, one estimation task is to infer some $(L_{xx'}, U_{xx'})$ such that  $L_{xx'} \leq \mathbb E[Y~|~do(X = x)] - \mathbb E[Y~|~do(X = x')] \leq U_{xx'}$.

This task has not received comparatively as much attention as the identifiable cases, despite promising early work in this area \citep{chickering96, Balke1997}. Lately the topic has sparked some interest in reinforcement learning \citep{kallus_zhou_2020, zhang2020},  algorithmic fairness \citep{wu2019,wu2019b}, and other settings \citep{gunsilius2019bounds, zhang2021, hu2021}. There has also been more interest recently in the problem of bounding causal effects from the perspective of sensitivity analysis \citep{freidling2022sensitivity, marmarelis2022bounding, jesson2022scalable}. \citet{xia2021_ncm} outline a procedure for bounding causal effects with neural networks, but restrict their focus to discrete, low-dimensional data. \citet{duarte2021} and \citet {zhang_poly2021} reduce the bounding problem to a polynomial program, but both assume that variables are discrete and \citet{zhang_poly2021} further assumes that variables are finite. In a recent paper, \citet{kilbertus2020class} propose a method for computing causal bounds in IV models with continuous treatments using gradient descent and Monte Carlo integration. Their procedure however is limited to univariate treatment settings (see \cref{app:limiations} for further limitations). 

In this work, we consider the problem of bounding causal effects for \emph{continuous} data with \emph{multivariate} treatments and auxiliary variables. Such models, based on variable \emph{clusters} \citep{anand:2022}, are particularly useful as they can help circumvent the need for excessive untestable assumptions in complex causal models. However, the setting of partial identification is conceptually tricky. The most direct way of tackling the problem of hidden confounding is by postulating distributions over black-box hidden variables \citep[e.g., the GAN-based setup of][]{hu2021}. Unfortunately, these distributions are not only completely uninterpretable, they also do not imply any testable implications to the observed marginals without further assumptions \citep{gunsilius2018testability}. This is true even in the highly constrained instrumental variable setup.

\begin{figure}[!t]
\begin{subfigure}{0.65\textwidth}
  \raisebox{0.89cm}{
  \hspace*{0.1cm}
  \begin{tikzpicture}
    \node[right] (L) at (-0.3, 1.8) {\textbf{a.}};
    \node[obs] (Z) at (0, 0) {$Z$};
    \node[obs] (X) at (1.2, 0) {$X$};
    \node[obs] (Y) at (2.4, 0) {$Y$};
    \node[latent] (U) at (1.2, 1) {$U$};
    \edge[black]{Z}{X};
    \edge[black]{X}{Y};
    \edge[->, bend right=30]{Z}{Y};
    \edge[black]{U}{Z};
    \edge[black]{U}{X};
    \edge[black]{U}{Y};
  \end{tikzpicture}%
  \hspace*{0.1cm}
  \begin{tikzpicture}
    \node[right] (L) at (-0.3, 1.8) {\textbf{b.}};
    \node[obs] (X) at (0, 0) {$X$};
    \node[obs] (M) at (1.2, 0) {$M$};
    \node[obs] (Y) at (2.4, 0) {$Y$};
    \node[latent] (U) at (1.2, 1) {$U$};
    \edge[black]{X}{M};
    \edge[black]{M}{Y};
    \edge[->, bend right=30]{X}{Y};
    \edge[black]{U}{X};
    \edge[black]{U}{M};
    \edge[black]{U}{Y};
  \end{tikzpicture}%
  \hspace*{0.45cm}
  \begin{tikzpicture}
    \node[right] (L) at (-0.3, 1.8) {\textbf{c.}};
    \node[obs] (Z) at (0, 0) {$Z$};
    \node[obs] (X) at (1.2, 0) {$X$};
    \node[obs] (Y) at (2.4, 0) {$Y$};
    \node[latent] (UZ) at (0, 1) {$U_Z$};
    \node[latent] (UX) at (1.2, 1) {$U_X$};
    \node[latent] (UY) at (2.4, 1) {$U_Y$};
    \edge[black]{Z}{X};
    \edge[black]{X}{Y};
    \edge[->, dashed, Firebrick3, bend right=30]{Z}{Y};
    \edge[black]{UZ}{Z};
    \edge[black]{UX}{X};
    \edge[black]{UY}{Y};
    \edge[<->, bend right=-30]{UZ}{UX}
    \edge[<->, bend right=-30]{UX}{UY}
    \edge[<->, Firebrick3, dashed, bend right=-45]{UZ}{UY}
  \end{tikzpicture}%
  \hspace*{0.3cm}
  \begin{tikzpicture}
    \node[right] (L) at (-0.3, 1.8) {\textbf{d.}};
    \node[obs] (X) at (0, 0) {$X$};
    \node[obs] (M) at (1.2, 0) {$M$};
    \node[obs] (Y) at (2.4, 0) {$Y$};
    \node[latent] (UX) at (0, 1) {$U_X$};
    \node[latent] (UM) at (1.2, 1) {$U_M$};
    \node[latent] (UY) at (2.4, 1) {$U_Y$};
    \edge[black]{X}{M};
    \edge[black]{M}{Y};
    \edge[->, dashed, Firebrick3, bend right=30]{X}{Y};
    \edge[black]{UX}{X};
    \edge[black]{UM}{M};
    \edge[black]{UY}{Y};
    \edge[<->, Firebrick3, dashed, bend right=-30]{UX}{UM}
    \edge[<->, bend right=-30]{UM}{UY}
    \edge[<->, bend right=-45]{UX}{UY}
  \end{tikzpicture}%
 }\\[-5mm]
 \end{subfigure}
 \vspace{-5mm}
 \caption{Graphs in a. and b. depict generic causal graphs for two classes of problems involving a treatment vector $X$ and an outcome vector $Y$: the first with pre-treatment causes $Z$, and the second with post-treatment mediators $M$, all confounded by a possibly infinite-dimensional vector of hidden common causes $U$. Graphs in c. and d. suggest partitioning $U$ into disjoint direct causes of the observed vectors, confounded by some black-box dependency structure (represented by mutual bi-directed edges). Dashed edges in red indicate submodels of interest obtained by removing those edges, submodels which can provide informative bounds on the causal relation between $X$ and $Y$.}
 \label{fig:coarse_models}
 \vspace{-2mm}
\end{figure}
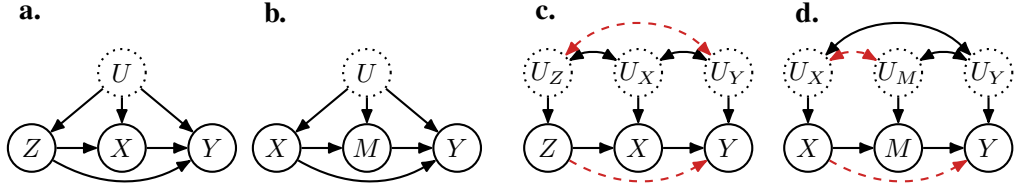

To address this, we make the following contributions: (1) We propose a generic, modular form of the effect bounding problem compatible with a wide range of graphical structures, function classes, optimization procedures, and distance measures; (2) We derive a customized method for \emph{clustered models}, where observed variables are separated into three vectors, treatment vector $X$, outcome vector $Y$, and a third vector of auxiliary variables, which are structurally constrained in the way they interact with $X$ and $Y$. Figure \ref{fig:coarse_models} provides an initial depiction of such classes. This not only covers a large array of practical problems where a complex causal graph is inapplicable but also, jointly with our optimization setup, \emph{removes distributional assumptions about unknowable hidden causes}. (3) We illustrate our method on synthetic and semi-synthetic datasets, where simulations confirm that our approach computes valid bounds even in complex settings that are challenging to optimize for.

\section{Setup}
\label{sec:setup}

We first present a convenient way of formulating structural causal models \citep[SCMs,][]{pearl2009causality} that is useful for the problem of partial identification, followed by a generic problem formulation for bounding unidentifiable causal estimands. In Section \ref{sec:method}, we then specify how to solve the practical clustered case in a way that minimizes distributional assumptions.

\subsection{SCMs and The Response Function Framework}
\label{sec:rf}
Let each variable $V \in \mathcal V$ be generated by a \emph{structural equation} $f_V(\mathrm{Pa}_V, U_V)$, where $\mathrm{Pa}_V$ are the direct causes of $V$ in $\mathcal V$ and $U_V$ is a potentially infinite-dimensional \emph{background} hidden vector of ``other causes'' of $V$. A generic stochastic process on the set of background variables $\{U_1, U_2, \dots, U_{|\mathcal V|}\}$ is assumed to exist.

Hence, specifying a causal model may require the joint distribution of potentially infinite-dimensional hidden confounders, along with the structural equations. To tackle this complexity, we make use of a \emph{response function} framework \citep{balke1994counterfactual, kilbertus2020class}, see also \cite[Section 3.4]{peters2017elements}.
Let $f_{V,u}(\cdot) := f_V(\cdot, u)$ be the response function for the \emph{fixed} value~$U_V\!=\!u$ of the background variable $U_V$. %
Function~$f_{V,u}$  encodes the direct causal effect of $\mathrm{Pa}_V$ on $V$ at realization $u$. A distribution over $U_V$ entails a joint distribution over functions~$f_{V,U_V}(\cdot) := f_V(\cdot, U_V)$ evaluated at inputs $\mathrm{Pa}_V$.%

The implication is that instead of explicitly modeling the distribution of background variables and, separately, the function space of the structural equations, we can directly encode a model via a distribution over response functions. This has interpretability and algorithmic advantages. To see this, consider the following $K_V$-dimensional function space for any given $V$,
\begin{equation}\label{eq:f_theta}
  \displaystyle
  \cF_V := \left\{f_{V, \boldsymbol{\theta}}(\cdot) := \sum_{k=1}^{K_V} \theta_{Vk} \psi_{Vk}(\cdot) \, \Big| \, \theta_{Vk} \in \bR \right\}\:.
\end{equation}
Instead of postulating distributional assumptions over background variables $U$, we directly postulate a distribution over parameter vector $\boldsymbol{\theta}_{V} := [\theta_{V_1}, \dots, \theta_{V_{K_V}}]$. Each basis function $\psi_{Vk}(\cdot)$ is a function of $\mathrm{Pa}_V$ only, i.e. $f_{V, \boldsymbol{\theta}}(\mathrm{Pa}_V) = \boldsymbol{\theta}_{V}^{\mathsf T} \boldsymbol{\psi}_{V}(\mathrm{Pa}_V)$. This is equivalent to assuming that the ``true'' structural equation has a finite (but arbitrarily large) factorized format 
$$f_{V, u}(\mathrm{Pa}_V) := \sum_{k = 1}^{K_V} \phi_{Vk}(u)\psi_{Vk}(\mathrm{Pa}_V),$$
 \noindent for some finite representation $\phi_{V\cdot}(\cdot)$ of the latent process $U_V$. This type of product representation, between two subsets of inputs to a structural equation, has been shown to be generic even in very complex treatment spaces \citep{kaddour:2021}, having numerous advantages as we shall see.

At first, we can assume a parameterized family of distributions over~$\cF$, the joint product function space over all of $\mathcal V$, via distributions over $\boldsymbol{\theta}_{V}$, denoted by $\{p_{\eta_V} \given \eta_V \in \bR^{D_V} \}$. There is, however, only so much data can tell us about the choice of $D_V$ and $K_V$. Without restrictions on the functional dependence between $V$ and~$U_V$, bounds are provably vacuous \citep{gunsilius2018testability,gunsilius2019bounds}.
As in general we have little information about the dimensionality and distribution of~$U_v$, we argue that it is more practical to work with the equivalent assumptions on the function space and distribution of response functions parameterized by $\boldsymbol{\theta}_{V}$. Specifically:
\begin{enumerate}
    \item  We can choose the function space $\cF_V$ to capture knowledge of the domain. For instance, we can inquire an expert about, ``all other things being equal'' (a fixed background variable $U_V$), how smooth a response function should be and what a plausible maximum number of inflection points  it could have. It may be plausible to exclude  having more than one inflection point, for instance, for the dose-response conditional effect of a drug dosage on a health outcome, given all other known and unknown factors \citep{gupta:2020}. Exploiting this knowledge can lead to tighter bounds;
    \item Depending on which information from the observed data will be used to constraint the function space in order to compute bounds, we may be able to get away with a partially specified model for $p_{\eta_v}$. As we shall see in the sequel, as the factorized formulation of (\ref{eq:f_theta}) puts all randomness in the generation of $V$ within the randomness of $\boldsymbol{\theta}_{V}$, it may be enough to specify $p_{\eta_V}$ only up to some of its moments \emph{while retaining a very flexible function space}.
\end{enumerate}

We can define each $\psi_{Vk}$ to itself have parameters, which can be fit along $\eta_V$. In most examples that follow, however, we will choose to keep the basis functions fixed to demonstrate in the experiments the trade-off between sources of causal knowledge, other than conditional independence constraints, and the informativeness of the bounds. See Appendix \ref{appendix:function_class} for more on adaptive basis functions. %

\subsection{A Causal Mathematical Program}

Assume we observe data from a causal model~$S^{\star}$, with density function $p_{S^\star}$ over continuous variables, including a (possibly multivariate) continuous treatment~$X \in \bR^p$ and outcome~$Y \in \bR$ among the observed variables. Let $o(\mathcal S^*)$ be a causal estimand of interest. For instance, using Pearl's \emph{do} notation \citep{pearl2009causality}, the functional $\E_{S^{\star}}[Y \given do(X=x^{\star})]$.
Our goal will be to minimize/maximize $o_{x^{\star}}(S^{\star})$ over all \emph{feasible} causal models $S$ among a (uncountable) model class $\cS$. Feasible here means that the causal model lies in the feasible region of the constrained optimization problem.
We characterize feasibility via two types of constraints. 
First, the missing edges of the assumed directed acyclic graph (DAG) encode conditional independencies between variables.
Further assumptions on the functional form of structural equations are required for non-vacuous solutions \citep{kilbertus2020class,gunsilius2018testability}.
We refer to the constraints implied by graphical and functional assumptions jointly as \emph{structural constraints}.
Second, we have observational data from $S^{\star}$. Let an estimate of the data distribution $p_{S^\star}$ be $\hat{p}$. Let $p_S$ be the distribution entailed by an admissible causal model~$S$. Then $p_S$ should replicate $\hat{p}$ as closely as possible. We measure this \emph{data constraint} via a distance function $\dist(p_S, \hat p)$ into $\bR^+$ that measures the (estimated) discrepancy between model $S \in \cS$ and the ground truth model $S^{\star}$ via observations.
Given this, we can formulate the following general problem setting for computing the minimum/maximum causal effect among all models $\cS$ that are $(\dist, \epsilon)$-compatible with observations:
\begin{align}
  \underset{\eta_\theta, w_\psi}{\text{min / max}}\quad
  & o(S)
  &&\obj
    \label{eq:optimization} \\
  \text{subject to}\quad
  & \dist(p_{S}, \hat{p}) \le \epsilon\:,
  && \cdata{}
    \label{eq:constraintdata} \\
  & \text{structural\ constraints,} \:
  && \cstruct{}
    \label{eq:constraintassumptions}
\end{align}
\noindent where: $\eta_\theta$ defines the joint distribution $p_\eta(\theta)$ over a finite-dimensional representation of the background variables in the construction of equations (\ref{eq:f_theta}) for each variable; $w_\psi$ parameterizes the respective basis functions, where applicable. The objective function {\tt [obj]} and constraints {\tt [c-data]} are well-defined as a function of $(\theta, \psi)$ and the distribution $p_\eta(\theta)$. The shape of the $p_\eta(\theta)$ and the structural equations (\ref{eq:f_theta}) should be such that {\tt [c-struct]} holds by construction.

One way to implement the above for $\mathcal{V} := \{V_1, \ldots, V_d\}$ is to define a deep generative model \citep[e.g.,][]{papamakarios2021} for the distribution over the set $\{\boldsymbol{\theta}_{1}, \ldots, \boldsymbol{\theta}_{d}\}$. Distance functions can in general be defined as $$\dist(p_S, \hat p) := ||T_S - \hat T||_p,$$
where $||\cdot||_p$ is a norm such as $L_2$ or $L_\infty$, and $T_\mathcal S$ is a (possibly infinite) vector of \emph{functionals} of $p_S$. Likewise, $\hat T$ is the (smoothed) empirical counterpart of this vector of functionals. One related alternative is to define this distance in terms of a set of maximum mean discrepancies \citep[MMD,][]{muandet:2017} at different choices of hyperparameters. 

\paragraph{Algorithm.}

The key challenges in implementing the method are two-fold. First, Monte Carlo evaluation of both {\tt [obj]} and {\tt [c-data]} will in general be necessary \citep[see, e.g.,][]{kilbertus2020class}. The result is a non-linear mathematical program with Monte Carlo evaluations, and constraints enforced by methods such as augmented Lagrangian optimization \citep{nocedal2006numerical}, which from now on we will refer to as \emph{stochastic causal programming} (SCP). Second, the parameterization of $p_S$ will need to enforce {\tt [c-struct]}. The corresponding graphical representation of a SCM is a acyclic mixed directed graphical model \citep[ADMG, see][for its Markovian properties]{richardson2003}, a graph with directed and bi-directed edges (as in Fig. \ref{fig:coarse_models}). The lack of directed edges can be easily enforced by the scope of each structural equation. The lack of bi-directed edges, however, may require a special treatment. A common trick is to introduce yet another layer of hidden variables according to the cliques of the bi-directed substructure of the ADMG, as done in many contexts by e.g. \cite{silva:07}, \cite{barber:09} and \cite{zhang_poly2021}. This implies a considerable computational burden. Furthermore, at least in the continuous case, the plausibility of the causal model may be compromised, as such a workflow invites imposing distributions over hidden variables in a mechanical way, without a proper discussion of what they entail in terms of causal assumptions.

\section{Approaches for the Clustered Case}
\label{sec:method}

The generic recipe for a SCP can be implemented by combining off-the-shelf components, as we have done, as described in \cref{app:general-scp}. This \emph{general SCP} is applicable to any ADMG structure with function spaces, as in (\ref{eq:f_theta}). However, the most interesting use cases should exploit the special structure of the domain. In what follows, we look at two important scenarios, found in many applications:

\begin{enumerate}
    \item The first comes from Figure \ref{fig:coarse_models}c., where we remove the dashed edges. As this is the same structure found in instrumental variable models, we call it the {\bf IV} family;
    \item The second comes from Figure \ref{fig:coarse_models}d., again where dashed red edges are removed. As it resembles the structure exploited by the (identifiable) front-door criterion of \cite{pearl2009causality} but for the confounding of $M$ and $Y$, we call this the {\bf leaky mediator} family.
\end{enumerate}

These families are meant to capture two major classes of applications: the first representing pre-treatment variables which explain part of the non-causal backdoor association between treatment and outcome; the second representing mediators which explain (part of) the causal effect between treatment and outcome. In both cases, allowing for the edge $Z \rightarrow Y$ or $X \rightarrow Y$ is trivial, but further constraints (such as a bound on the derivative of the structural equation of $Y$ with respect to the added parent) will be necessary to avoid vacuous bounds \citep{ramsahai:2012}. We do acknowledge that combinations of the IV and leaky mediation scenarios appear in longitudinal studies, but we leave a specialized treatment of such a case to future work.
 
\subsection{Parameterization}
\label{subsec:cstruct}

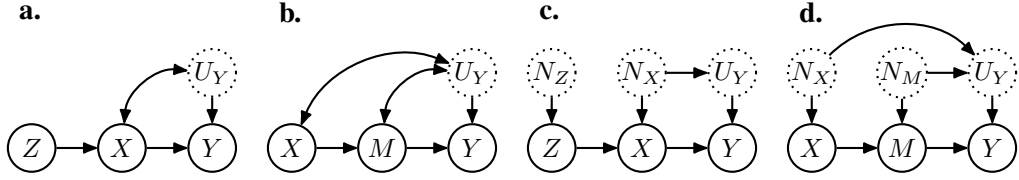
\begin{figure}[!t]
\begin{subfigure}{0.65\textwidth}
  \raisebox{0.89cm}{
  \hspace*{0.1cm}
  \begin{tikzpicture}
    \node[right] (L) at (-0.3, 1.8) {\textbf{a.}};
    \node[obs] (Z) at (0, 0) {$Z$};
    \node[obs] (X) at (1.2, 0) {$X$};
    \node[obs] (Y) at (2.4, 0) {$Y$};
    \node[latent] (UY) at (2.4, 1) {$U_Y$};
    \edge[black]{Z}{X};
    \edge[black]{X}{Y};
    \edge[black]{UY}{Y};;
    \edge[<->, bend right=-45]{X}{UY}
  \end{tikzpicture}%
  \hspace*{0.3cm}
  \begin{tikzpicture}
    \node[right] (L) at (-0.3, 1.8) {\textbf{b.}};
    \node[obs] (X) at (0, 0) {$X$};
    \node[obs] (M) at (1.2, 0) {$M$};
    \node[obs] (Y) at (2.4, 0) {$Y$};
    \node[latent] (UY) at (2.4, 1) {$U_Y$};
    \edge[black]{X}{M};
    \edge[black]{M}{Y};
    \edge[black]{UY}{Y};
    \edge[<->, bend right=-45]{X}{UY}
    \edge[<->, bend right=-45]{M}{UY}
  \end{tikzpicture}%
  \hspace*{0.3cm}
  \begin{tikzpicture}
    \node[right] (L) at (-0.3, 1.8) {\textbf{c.}};
    \node[obs] (Z) at (0, 0) {$Z$};
    \node[obs] (X) at (1.2, 0) {$X$};
    \node[obs] (Y) at (2.4, 0) {$Y$};
    \node[latent] (NZ) at (0, 1) {$N_Z$};
    \node[latent] (NX) at (1.2, 1) {$N_X$};
    \node[latent] (UY) at (2.4, 1) {$U_Y$};
    \edge[black]{Z}{X};
    \edge[black]{X}{Y};
    \edge[black]{NZ}{Z};
    \edge[black]{NX}{X};
    \edge[black]{UY}{Y};
    \edge[black]{NX}{UY};
  \end{tikzpicture}%
  \hspace*{0.3cm}
  \begin{tikzpicture}
    \node[right] (L) at (-0.3, 1.8) {\textbf{d.}};
    \node[obs] (X) at (0, 0) {$X$};
    \node[obs] (M) at (1.2, 0) {$M$};
    \node[obs] (Y) at (2.4, 0) {$Y$};
    \node[latent] (NX) at (0, 1) {$N_X$};
    \node[latent] (NM) at (1.2, 1) {$N_M$};
    \node[latent] (UY) at (2.4, 1) {$U_Y$};
    \edge[black]{X}{M};
    \edge[black]{M}{Y};
    \edge[black]{NX}{X};
    \edge[black]{NM}{M};
    \edge[black]{UY}{Y};
    \edge[black]{NM}{UY}
    \edge[->, bend right=-45]{NX}{UY}
  \end{tikzpicture}%
  \hspace*{0.1cm}
 }\\[-5mm]
 \end{subfigure}
 \vspace{-5mm}
 \caption{a.\ and b.\ show the causal structure of the main pre-/post-treatment templates, ignoring the background variables that will be irrelevant for our tasks. In c.\ and d. we introduce (non-structural) hidden causes $N_{\cdot}$, which will allow for a parameterization of hidden confounding. Please notice that graphs in c.\ and d. should \emph{not} be interpreted as representing causal relations between $N.$ and their children: $N.$ are just data-augmentation devices to build a distribution, and the graphs here just represent standard probabilistic Markov conditions.}
 \label{fig:dag_model}
 \vspace{-2mm}
\end{figure}

The main relevant common feature found in the IV and leaky mediator is the way we can deal with the marginal independence model implied by the lack of bi-directed edges $U_Z \leftrightarrow U_Y$ and $U_X \leftrightarrow U_M$ \citep{verma1990causal}. The main idea comes from observing that the only structural confounding of interest is the association between $U_Y$ and the ancestors of $Y$. Figures \ref{fig:dag_model}a. and b. reconceptualize the independence model of the IV\footnote{The graphical criteria for an IV remains the same regardless of the existence of edge $U_Z \leftrightarrow U_X$ \citep{pearl2009causality}.} and leaky mediator cases in terms of treatment $X$, outcome $Y$, structural background variable $U_Y$, and the corresponding auxiliary variable ($Z$ or $M$). 

Following the standard idea of transforming cliques of bi-directed edges by adding a latent variable vector per clique in a DAG model \citep[e.g.][]{silva:07}, we can reconceptualize Figures \ref{fig:dag_model}a. and b. as the DAG models in Figures \ref{fig:dag_model} c. and d. Hidden variables $N_Z$ and $N_M$ are \emph{not} structural, meaning that they do not represent background variables in a SCM. We call them the \emph{non-structural} hidden variables. Indeed, such models do not need to assume $Z$ and $M$ are manipulable at all as long as they encode the causal invariances that restrict the model space, such as $M(x) \indep X$ and $Y(x) \indep Z$, where $M(x)$ and $Y(x)$ are the respective potential outcomes, under an intervention on $X$, in the respective models. Non-independence invariances arise in the leaky mediator model even when $Y(m)$ is not defined \citep[see][for the reasoning in the case where the front-door criterion holds]{dawid:2022}. For simplicity of presentation, we shall assume that the equation for $Y$ in the leaky mediator is structural with $M$ as a causal parent such that the constraint $Y(m, x) = Y(m, x')$ is well-defined, but we will not make use of a structural equation for $M$. %

Using the parameterization from \cref{eq:f_theta}, we may now reformulate structural causal claims in terms of the response function representation of vector $U_V$ as vector $\boldsymbol{\theta}_V$. For instance, ignorability between $V$ and its observable parents in the ADMG can be cast as $\boldsymbol{\theta}_V \indep Pa_V$.

\paragraph{Structural hidden variables.} Following the Markovian structure of the IV and leaky mediator, it suffices to represent the distribution
\begin{equation}\label{eq:thetadistr}
    \boldsymbol{\theta}_Y \given N_{\mathrm{Pa}_Y} \sim p_{\eta_Y}\bigl(\cdot; \mu_{\eta_0}(N_{\mathrm{Pa}_Y}), \Sigma_{\eta_1}(N_{\mathrm{Pa}_Y})\bigr),
\end{equation}
where $N_{\mathrm{Pa}_Y}$ are the non-structural hidden variable parents (in terms of the independence structure of the graphical model) of $U_Y$ in the SCM. This distribution is defined up to mean and covariance functions $\mu_{\eta_0}(\cdot)$ and $\Sigma_{\eta_1}(\cdot)$, defined by our optimization parameters~$\eta = (\eta_0, \eta_1)$. Other than that, we make \emph{no further assumptions} about the shape of $p_{\eta_Y}$, as they are unnecessary for the inference problems we set up to solve in the next section. To ensure $\Sigma_{\eta_1}(N)$ is a valid covariance matrix, we represent it in terms of its Cholesky factor~$L$ and add a small constant $\Omega$ to the diagonal $\Sigma_{\eta_1} = L^{\top} L + \Omega \B{1}$.
In our implementation of Section \ref{sec:experiments}, we use flexible function approximators for $\mu_{\eta_0}$ and $\Sigma_{\eta_1}$ given by feedforward neural networks.

\paragraph{Non-structural hidden variables.} 

Unlike black-box approaches, including the one we presented in the previous section and related approaches such as \cite{hu2021, zhang_poly2021}, in our clustered case there is no need to waste computational resources and making further model assumptions by defining a model for $N_V$, the non-structural hidden variables. Instead, define 
$$V = h_{\mathrm{Pa}_V}(N_V)$$ 
as the model of a cluster $V$ with parent clusters $\mathrm{Pa}_V$ and respective non-structural hidden parent $N_V$ in a way that $h_{\mathrm{Pa}_V}(\cdot)$ is invertible in $N_V$. That is,
\begin{equation}
    N_V = h^{-1}_{\mathrm{Pa}_V}(V).
\end{equation}
When $V$ is one dimensional, this can be done using the usual construction based on inverse cumulative distribution functions (cdf). That is, if $F_{\mathrm{Pa}_V}(\cdot)$ is the cdf of $V$ given $\mathrm{Pa}_V$, then we can define it as $V = F^{-1}_{\mathrm{Pa}_V}(U)$ for a uniform in $(0, 1)$ random variable $U$ (recall we are assuming that all variables are continuous). This means we can define our $N_V$ as $U$ itself. Moreover, given $(V, \mathrm{Pa}_V)$ $U = N_V = F_{\mathrm{Pa}_V}(V)$ allows for actually \emph{observing} $N_V$ within any data point.

When $V$ is multivariate, all sorts of approaches can be used. A convenient one is conditional normalizing flows composed of invertible and differentiable (i.e., diffeomorphic) transformations \citep{papamakarios2021}. They are flexible candidates that we use in our experiments. We provide all details of our specific implementation in \cref{app:implementation}. The fact that we can deterministically infer the non-structural hidden variables will vastly simplify our algorithm, as we shall see next. %

\subsection{Objective functions and enforcing \cdata{}}
\label{subsec:data}

Now that we described how the model space is defined, we finish the specification of the causal program by constructing the objective function and data constraints.

\paragraph{Objective functions.} The objective function we will assess in our experiments is the \emph{expected outcome under intervention},
\begin{equation}
    o_{x^\star} := \mathbb E[Y(x^\star)] = \mathbb E[Y~|~do(X = x^\star)],
\end{equation}
\noindent for a particular value $x^\star$ of interest. This can be used to generate pointwise bounds over a dose-response curve on a grid of treatment values, for instance. Average treatment effects (ATEs) easily follow, and pre-treatment variables can be trivially added for deriving conditional average treatment effects (CATEs), as similarly discussed by \citep{kilbertus2020class}. The function $o_{x^\star}$ is particularly easy to compute in the case without mediators, where $$o_{x^\star} = \psi_Y(x^{\star})^\top \E_{N_X}[\mu_{\eta_0}(N_X)].$$ The expectation is over the marginal distribution of $N_X$. Defining the model for $X$ as a normalizing flow with independent Gaussian seeds $N_X$ means that this objective function can be easily computed by a Monte Carlo approximation of the integral over $N_X$.

When mediator $M$ is at play, we need also to integrate out $M(x^\star)$. As $M(x^\star)$ is deterministically given by $N_M$, and $o_{x^\star}$ is still a linear function of $\theta_Y$, this can be much simplified as
\[
\begin{array}{rcl}
\E[Y~|~do(X = x^\star)] &=& \E_{M(x^\star), \theta_Y}[\psi_Y(M(x^\star))^\top \theta_Y]\\
&=&\E_{N_X, N_M}\E_{M(x^\star), \theta_Y}[\psi_Y(M(x^\star))^\top \theta_Y~|~N_X, N_M]]\\
&=&\E_{N_X, N_M}[\psi_Y(h_{x^\star}(N_M))^\top \mu_{\eta_0}(N_X, N_M)].\\
\end{array}
\]

In case of cross-world counterfactual quantities, $\mathbb V[Y(x) - Y(x')]$ is an example of estimand which is useful as a summary of the joint distribution of counterfactuals and can also be easily calculated. We will not experimentally pursue cross-world estimands further in our manuscript.

\paragraph{Encoding the \cdata{} constraints.}

Classical methods such as \cite{balke1994counterfactual} match the entire likelihood function implied by the model against the maximum likelihood estimator of the observational distribution. With continuous realizations, we cannot match entire likelihoods. However, matching only at a particular set of realizations will not invalidate the bounds.

Let $W$ be the auxiliary variable $Z$, or $M$, depending on the model. From the factorization $p(W, X, Y) = p(Y \given X, W)\, p(X, W)$, the model for $p(X, W)$ can be learned independently of causal parameters. This means, for instance, making $N_Z = Z$ and fitting separately a normalizing flow for $X$ as a function of $Z$ and $N_X$ prior to any causal effect bounding. 

A key feature of the clustered method is then only matching functionals of $p(Y \given X, W)$. This matching can be conveniently formulated as matching expectations of the two distributions when transformed by a set of dictionary functions $\{\phi_l: \cY \to \bR \}_{l=1}^L$.
That is, we want some norm of
\begin{equation}\label{eq:constraintentry}
  \nu_l(x,w) = \E_{\hat{p}(Y\given x, w)}[\phi_l(Y)] - \E_{p_{\eta_Y}(\theta_Y \given x, w)}[\phi_l(f_{\theta_Y}(x))]  \: 
\end{equation}
defined as $\|\nu_l(\cdot, \cdot)\|$ at a representative, finite set of points in the support of $p(X, W)$.
Arguably, the most representative set is a uniformly random subsample of the observed data~$\cD$ of size~$D$. In our experiments, we specifically consider the entry-wise sup-norm~$\|\cdot\|_{\infty,\infty}$ and the entry-wise 2-norm~$\|\cdot\|_{2,2}$.
For~$\|\cdot\|_{\infty, \infty}$ we require the absolute value of each entry to be small, i.e., practically have to enforce $D\cdot L$ constraints, whereas for~$\|\cdot\|_{2,2}$ we get away with a single overall constraint.

Choosing $L = 2$ with $\phi_1(s) := s$ and $\phi_2(s) := s^2$ will allow us to calculate constraints in \emph{closed form}, while keeping the premise that $p_{\eta_Y}(\theta_Y)$ can be defined only up to (conditional) first and second moments $\mu_{\eta_0}(N_{\mathrm{Pa}_Y})$ and $\mu_{\eta_1}(N_{\mathrm{Pa}_Y})$. This task is more computationally efficient and use weaker assumptions than modeling the complete joint distribution $p(X, Y, W)$, as required, for example, by generative methods such as GANs \citep{hu2021}.
For instance, in the IV case,
\begin{equation}
\begin{array}{lclcl}
 \rhs_{1,j}(\eta) &:=& \E[f_{\theta_Y}(X_j)~|~x_j, z_j] &=& \psi(x_j)^\top \mu_{\eta_0}(n_j),
 \\
 \rhs_{2,j}(\eta) &:=& \E[f_{\theta_Y}(X_j)^2~|~x_j, z_j] &=& \psi(x_j)^\top \bigl(\Sigma_{\eta_1}(n_j) + \mu_{\eta_0}(n_j) \mu_{\eta_0}(n_j)^\top \bigr)  \psi(x_j)\:,
\end{array}
\label{eq:rhs-form}
\end{equation}
where $n_j := h_{z_j}^{-1}(x_j)$, for each $j \in 1, 2, \dots, D$.

\begin{figure}[t]
\centering
  \includegraphics[width=0.43\linewidth]{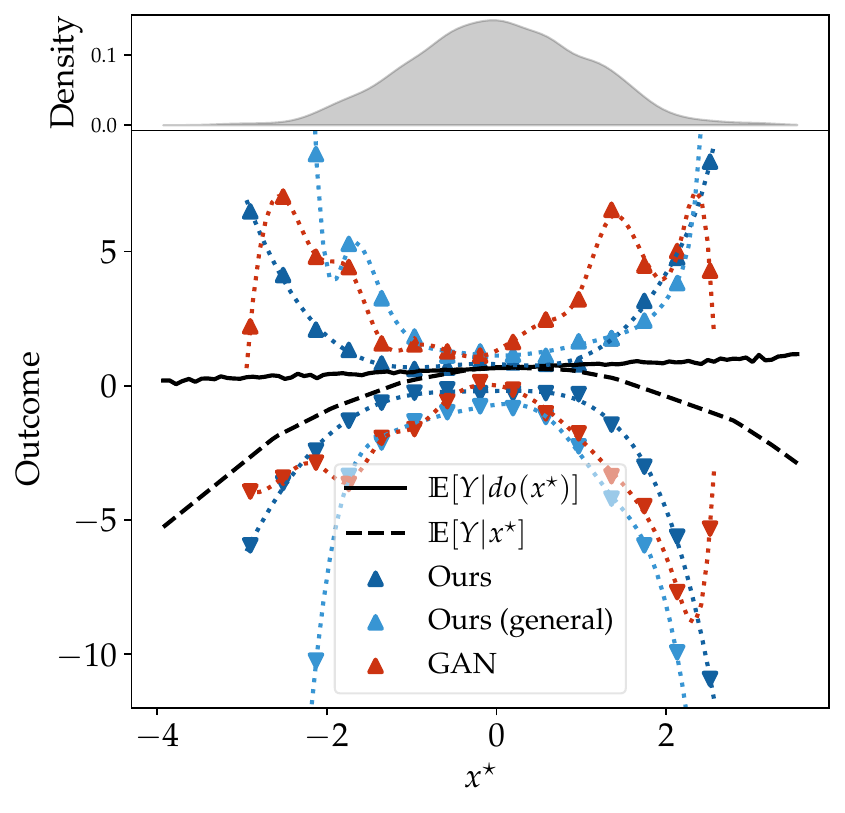}
  \hspace{1cm}
  \includegraphics[width=0.43\linewidth]{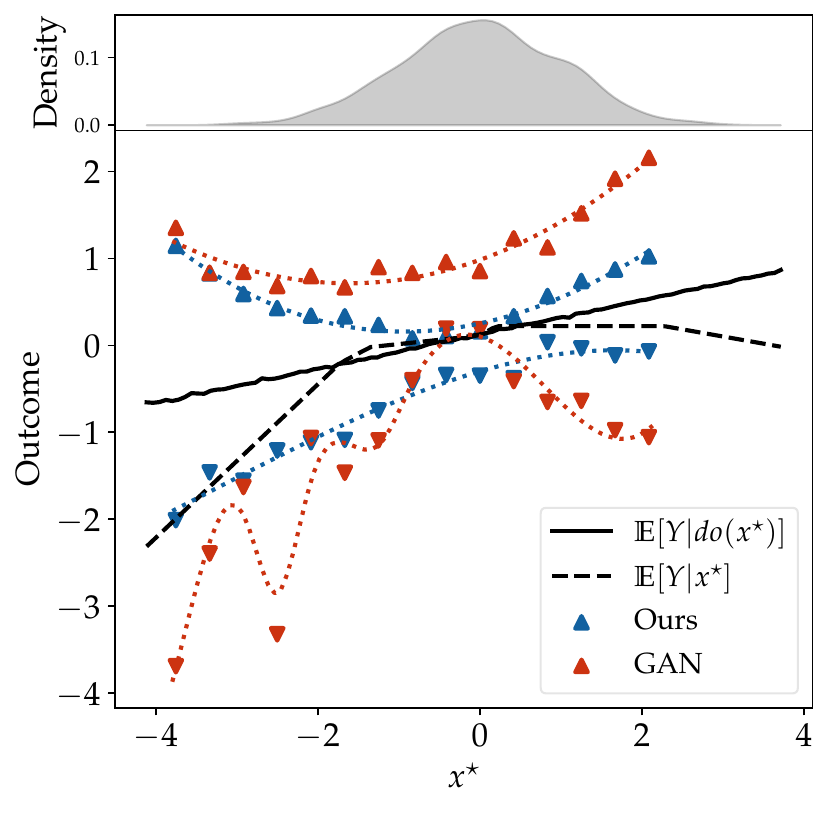}\\
  \includegraphics[width=0.43\linewidth]{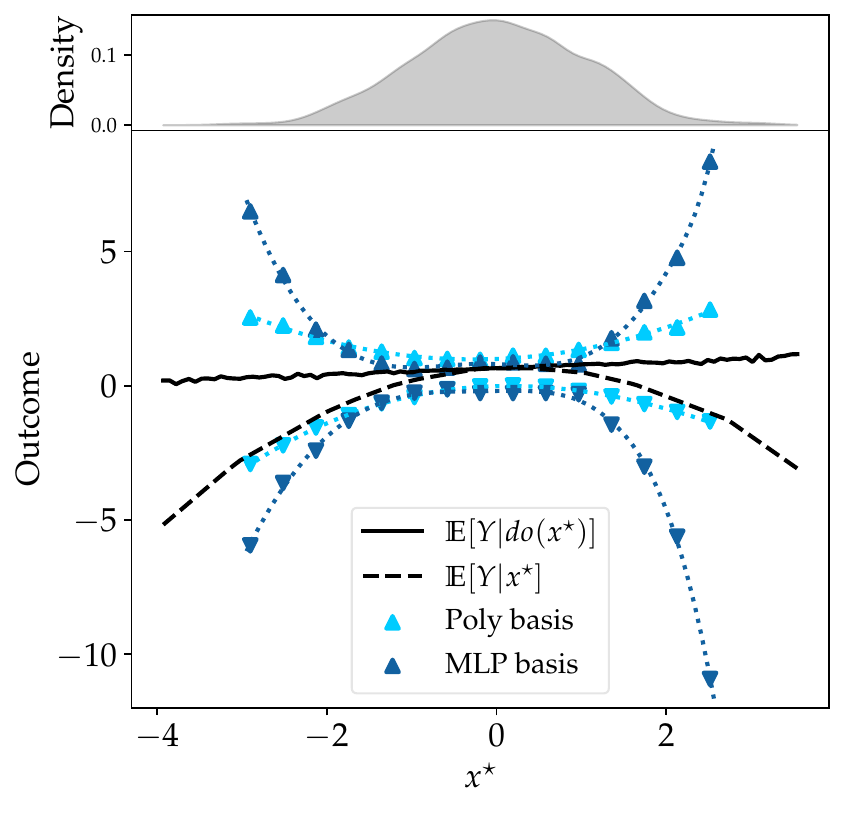}
  \hspace{1cm}
  \includegraphics[width=0.43\linewidth]{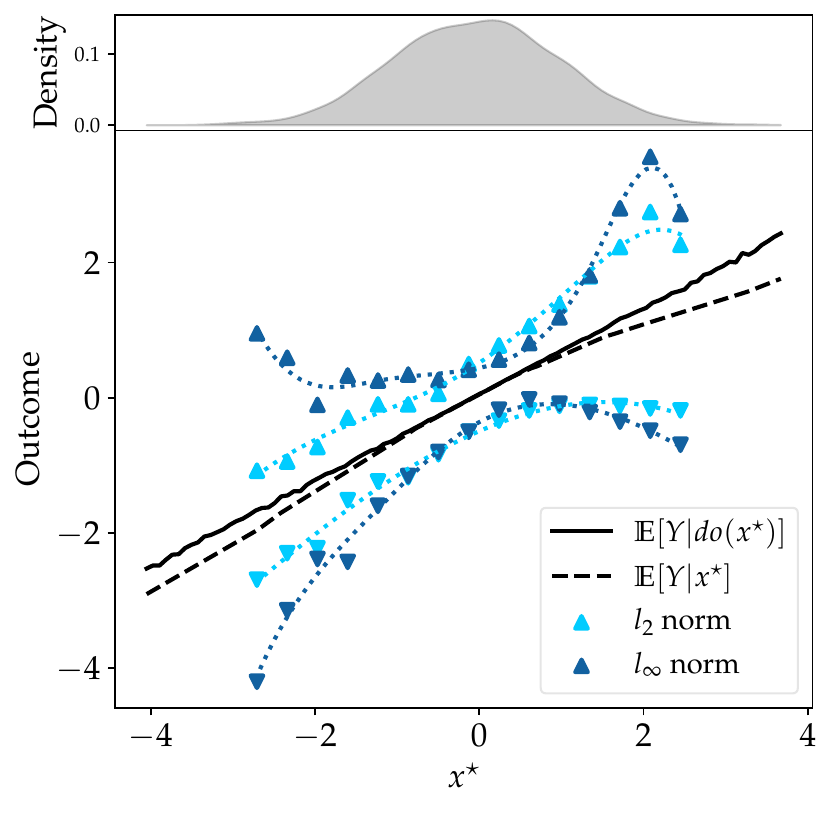}
  
  \caption{\textbf{Top Left} (IV-lin-2d-strong; GAN comparison) Our general SCP (\cref{app:general-scp} gives bounds comparable to the GAN method. The clustered SCP outperforms both the GAN and general SCP in terms of tightness of bounds and stability. \textbf{Top Right} (LM-lin-2d-strong; leaky mediator setting) We get reliable bounds also in the leaky mediator setting compared to the GAN framework. \textbf{Bottom Left} (IV-lin-2d-strong; comparing response functions) As expected, neural response functions give wider bounds than a linear polynomial basis because of being less expressive. \textbf{Bottom Right} (IV-lin-2d-weak; comparing norms) Our framework easily allows for different data-matching criteria ($\dist$). The choices shown here ($\|\cdot\|_{\infty,\infty}$ and $\|\cdot\|_{2,2}$) yield comparable and consistent bounds.}
  \label{fig:main-comparison} \vspace{-6mm}
\end{figure} 
\subsection{Solving the optimization}\label{subsec:optimization}
Taking all the steps from previous sections together, we have $\nu_{l,j} = \hat{\phi}_l(x_j, z_j) - \rhs_{l,j}(\eta)$ and ultimately arrive at the following non-convex optimization problem with non-convex constraints
\begin{align}
  \underset{\eta_{\theta_Y}, w_{\psi_Y}}{\text{min / max}}\;
  & o_{x^{\star}}(\eta)]
  &&\obj  \label{eq:constrainedfinal} \\
  \text{subject to}\;
  & \| \nu \| \le \epsilon
  && \cdata{}\nonumber
\end{align}
for a choice of norm $\|\cdot\|$ and where $\nu$ is a stacked vector of the left-hand side of all $\nu_l$-induced constraints. For this problem, the augmented Lagrangian method with inequality constraints is a natural choice \citep[Sect.~17.4]{nocedal2006numerical}, along with a Monte Carlo evaluation of gradients of the objective function (recall that $\|\nu\|$ and its gradients can be done in closed form). A high-level description and details are in \cref{app:implementation}.

\section{Experiments}
\label{sec:experiments}
Since ground truth causal effects must be known to properly evaluate the validity of our bounds, we make use of synthetic and semi-synthetic datasets. We show experiments for identifiable settings as well as for other datasets and constraint formulations in \cref{app:more-experiments}. All plots for our method are using the clustered formulation (\cref{sec:method}) unless specified otherwise.\footnote{Code for the experiments is available at \url{https://github.com/kirtanp/SCP_bounds}}
\subsection{Treatment choices}
\label{subsec:xstar-choice}
When visualizing results for multidimensional treatments~$X$, we vary the interventional values~$x^{\star}$ along a single treatment dimension, keeping the remaining components at fixed values.
While this allows us to show continuous treatment effect curves, we note our method can compute bounds for any multidimensional intervention $do(X=x^{\star})$.
Specifically, we vary the first component of $X$ and fix the values of the other components to their empirical marginal means.
For each figure, we include a kernel density estimate of the marginal distribution of the empirically observed treatments to distinguish ``data poor'' from ``data rich'' regions. In ``data poor'' regions (towards the tails of the empirical distribution), we expect our bounds to become looser, as less information about the data-matching constraints is available.

\subsection{Baselines}
We compare our clustered SCP method to a general SCP implementation (\cref{app:general-scp}),  na\"ive regression and the method of \citet{hu2021}, which is a typical representative of black-box methods for continuous treatments. 
\begin{itemize}[leftmargin=*,topsep=0pt,itemsep=0pt]
 \item \textbf{General SCP implementation.} We provide an implementation of the general formulation of SCP. The idea is that each variable is a flow conditioned on its parents. The structural constraints are enforced by the fact that all variables within each c-component share the same noise distribution, and noise distributions for different c-components are independent. Details are described in \cref{app:general-scp}.
 \item \textbf{Regression with MLP.}
We na\"ively fit an MLP with quadratic loss to predict outcome~$Y$ from the multidimensional treatment~$X$, a modeling approach that assumes no confounding at all.
 \item \textbf{GAN framework.}
\citet{hu2021} parameterize the causal model (i.e., its exogenous random variables and structural equations) with neural networks and apply the adversarial learning framework to search the parameter space. We adjust the model in their code to our examples, but otherwise leave hyperparameter choices and convergence criteria untouched. %
\end{itemize}

\subsection{Implementation choices}
\label{subsec:implementation}

\xhdr{Choice of response functions}
As described in \cref{eq:f_theta}, we choose linear combinations of non-linear basis functions $\{\psi_k\}_{k\in [K]}$.
For our experiments we mostly work with a set of pre-trained $K$ \emph{neural basis functions}, obtained from the last hidden-layer activations of an MLP fit to the observed data $\{(x_i, y_i)\}_{i\in [n]}$, as well as (multivariate) polynomials up to a fixed degree (see \cref{app:response} for details). For two-dimensional treatments, we use $K=6$ and $K=3$ for polynomial and neural basis functions, respectively. For three-dimensional treatments, we use $K=10$ for both polynomial and neural basis functions. This allows up to quadratic terms in the polynomial basis for both two- and three-dimensional treatments. We also show experiments with adaptive basis functions, with $\psi_k$ is $\tanh(w_{0k} + w_{1k}x)$. The bases in this case are therefore not fixed, but rather adapt to the problem at hand through the learned weights. More details are provided in \cref{app:response}.

\looseness=-1
For our method, we individually compute \textbf{lower} ($\bigtriangledown$) and \textbf{upper} ($\triangle$) bounds at multiple values $x^{\star} \in \bR$ for one dimension of the treatment and show how it compares to the \textbf{true causal effect} (\trueline{}) and \textbf{na\"ive regression} (\regressline{}).
The lines shown for \textbf{lower} and \textbf{upper} (\boundline{}) bounds are univariate cubic splines fit to the bounds for individual $x^{\star}$-grid values.
We use $n=10,000$ i.i.d.\ sampled data points for each experiment and subsample $M=100$ data points uniformly at random for the data constraints \cdata{}.
For the $X\given Z$ model in the IV setting, we use a quadratic conditional spline normalizing flow.
Details of the implementation, including hyperparameters, are described in \cref{app:implementation}.

\xhdr{Getting final bound estimates} For each $x^{\star}$, we run 5 optimizations with different seeds each for the lower and upper bounds. For the final upper bound estimate, we take the maximum of the 5 upper bound estimates we get, and for the final lower bound estimate, we take the minimum of the 5 lower bound estimates we get. We follow the same process for the GAN bounds.

\subsection{Results}\label{sec:results}
A full glossary of datasets can be found in \cref{app:data}, along with exact structural equations for each setting. While the naming of the datasets is intuitive, \cref{fig:app-naming-logic} provides a short description of the naming logic. We now describe the results obtained for selected datasets, with more results and details in \cref{app:more-experiments}.

\xhdr{IV-lin-2d-strong}
This dataset is simulated from an IV setting, where the true effect $Y \given do(X)$ is linear in a two-dimensional treatment $X$. The na\"ive regression $\E[Y\given X]$ differs substantially from the true effect, indicating strong confounding.

\emph{SCP is more stable than GAN.} \cref{fig:main-comparison} (top right) compares both the clustered and the general implementation to the GAN framework baseline. %
Our general SCP gives bounds comparable to the GAN method, though with fewer sharp jumps and instabilities. The clustered SCP on the other hand simplifies constraints through structural encoding and avoids full density estimation, which is reflected in the increased stability over both the general SCP and the GAN methods. Indeed, \cref{fig:main-comparison} (top right) shows that despite the flexible neural basis functions, our clustered approach can yield tighter bounds and avoid the instabilities observed in the GAN approach when we move towards the tails of the observational distribution.

While the tightness of bounds across different models with different assumptions should \emph{not} be an indicative of quality of a solution, the point here is to illustrate that using only a few moments of the distribution does not trivialize the resulting interval when contrasted against methods that use full distributions (GAN), and the bounds remain consistently valid. Furthermore, the stability of the bounds is especially encouraging when we remember that each upper and lower bound for each intervention value $x^*$ is a separate optimization problem.

\emph{The flexibility of the response functions affects the tightness of the bounds.} We continue by assessing the effect of the flexibility of response functions on our bounds in \cref{fig:main-comparison} (bottom left). Choosing less flexible basis functions (polynomials) yields tighter bounds, highlighting the flexibility of our approach in obtaining more informative bounds when more restrictive assumptions are made. Our method can also accommodate alternative constraint formulations and slack parameters. We show experiments demonstrating this flexibility in \cref{app:more-experiments}.
While sometimes loose, especially for flexible basis functions (MLP) and in ``data poor'' regions towards the tails of the empirical distribution of observed treatments, our bounds contain the true causal effect for all $x^{\star}$. This validity holds for all other experiments as well.

\xhdr{IV-lin-2d-weak} This is an IV setting with weak confounding. In \cref{fig:main-comparison} (top left), we show how our bounds behave under different choices of $\dist$, namely under the entry-wise $\|\cdot\|_{\infty, \infty}$ norm (which results in $D\cdot L$ constraints in the augmented Lagrangian) versus the entry-wise $\|\cdot\|_{2,2}$ norm (yielding only a single constraint). The obtained bounds are compatible and comparable, indicating relatively mild effects of the choice of the data-matching criterion $\dist$.

\xhdr{LM-lin-2d-strong} \Cref{fig:main-comparison} (top right) shows results for data from a leaky mediator setting, where the true effect is a linear function of the mediator $M$, and the treatment $X$ is two-dimensional. Confounding between the mediator and the effect is relatively strong.
These results corroborate our findings regarding stability and reliability of our bounds compared to the GAN framework from the IV setting, despite using the neural basis function. Again, additional results for different structural assumptions, treatment dimensions, and confounding strength are presented in \cref{app:more-experiments}.

\section{Discussion}\label{sec:limitations}
\xhdr{Limitations} After parameterization, the optimization problem in \cref{eq:optimization} will generally result in a \emph{non-convex objective with non-convex constraints}.
Therefore, our proposed gradient-based local optimization may not converge to a global optimum, possibly rendering our bounds overly tight. Empirically, we do not observe evidence of consistently getting stuck in bad local optima. Because we optimize both bounds individually for each value of $x^{\star}$, our method may be computationally expensive when the intervention space is very large. However, it is well-suited for scenarios where we want reliable bounds on a well-defined set of plausible interventions.
Finally, we have not accounted for the uncertainty of our bounds. Confidence or credible intervals for both extrema can help practitioners evaluate the reliability of causal inferences, and are an interesting direction for future work.

\xhdr{Conclusion} Causal modeling inevitably involves a trade-off between the strength of input assumptions and the specificity of resulting inferences. If assumptions do not correspond to reality, the inference that follows is unwarranted. The role of a method is to provide the sandbox in which assumptions can be transparently expressed and informative constraints, where available, are not wasted. Our framework allows for the practitioner to focus more on the coarse structure of the problem and assumptions about the function space, and to deliver results with minimal assumptions about non-causal aspects of the model, while also having the flexibility of making stronger assumptions when appropriate. See Appendix \ref{appendix:function_class} for further discussion. 

We have introduced a stochastic causal program for bounding treatment effects in partially identifiable settings. We showed how major computational simplifications and weakening of assumptions can be achieved in practical cases where variables can be clustered in pre- and post-treatment layers. Our approach does not rely on the typical assumptions of linearity, monotonicity, or additivity, nor on distributional assumptions about unknowable hidden common causes. Experiments on synthetic and semi-synthetic datasets demonstrate that our method produces valid and informative bounds in a wide range of settings, including with continuous and multidimensional instruments, mediators, treatments, and outcomes. 

\section*{Acknowledgments}
JZ was supported by UKRI grant EP/S021566/1. RS and DW were partially supported by the ONR grant 62909-19-1-2096. RS also acknowledges funding via the EPSRC Open Fellowship grant EP/W024330/1. We would also like to thank the anonymous reviewers for their insightful feedback, which helped us to strengthen our manuscript.
\bibliography{bib}

\clearpage
\appendix

\section{On the Choice of Function Space}
\label{appendix:function_class}

The function space in Eq. \eqref{eq:f_theta} is, at a high-level view, a standard finite basis expansion with or without adaptive bases. We chose this particular function space template as it is (i) general, (ii) computationally advantageous and (iii) scientifically advantageous. 

It is {\bf general}, as it can arbitrarily approximate any practical function of interest (keeping in mind that allowing for an unbounded number of discontinuities will result in vacuous bounds \citep{gunsilius2018testability}), particularly when we are dealing with a small number of causal parents. As a matter of fact, \citep{gunsilius2019bounds} uses as basis functions a truncated wavelet expansion. We could likewise use the dictionary of regression spline basis functions, with splits decided by any rule of interest, including uniform grids or those based on the training points.

As a matter of fact, no computable estimator of nonparametric functions is infinite dimensional at any given sample size. In the case of kernel machines, for instance, the celebrated Representer Theorem of \citep{scholkopf_rt:2001} shows how any resulting estimate will be a function that is parameterized explicitly by the data (and such  will be the case for the broad class of linear smoothers \citep[Ch. 5]{wasserman:2006}). As our causal models are \emph{not} fit to the data directly, but to estimated functionals of the observed distribution, which we use as constraints in a mathematical program, we can directly parameterize it in terms of a finite basis expansion. 

Moreover, there is nothing stopping us from making each element $\psi_i$ of the basis functions parameterized by, say, a linear transformation of the function inputs followed by some non-linearity (effectively, a hidden unit in a feedforward neural network). An entire neural network could also be used for each $\psi_i$, so that the function space becomes a deep model, with the distribution over the function space confined only to randomness on the $\theta$ parameters. Nothing in the main algorithm changes, except for extra terms added to the overall gradient.

This leads to the fact that our function space is {\bf computationally advantageous}, particularly when we can eliminate bidirected edges using ideas from \citet{drton2008}, combined with the decision to use only the first two moments of target causal factors. In this case, all constraints can be computed in closed-form, and we can perform a relatively simple Monte Carlo approximation for the objective function. As a matter of fact, we did not have to make any distributional assumptions about $\theta$ at all, except for the existence of its first and second moments. There is much to appreciate in the generic blackbox of \citet{hu2021}, but solving full density estimation problems and requiring a complex nesting of multiple optimization and simulation steps may not be the preferred way to go in some real-world applications.

Moreover, response functions only need to be structural with respect to its manipulable inputs. If we want to model, say, $\mathbb E[Y~|~do(x), w]$ for some pre-treatment covariates $W$, there is no need at all to model $W$ structurally. The causal model can be defined in terms of the response function $f_{(w, u_Y)}(x)$, where $W$ is just another source of background variability (though observable, unlike $U_Y$). This means that the \emph{distribution} of $\theta$ will be parameterized as a black-box function that includes $w$ as inputs, while the basis function expansion can remain flexible, and informative, in the small dimensional space of $X$.

Finally, our setup is {\bf scientifically advantageous}. A na\"{i}ve view of causal inference postulates that since functional constraints are untestable assumptions that will imply invalid bounds if misspecified, then we must, on purpose, impoverish the language in which we allow causal models to be described by removing this choice. This line of reasoning is incoherent on many accounts. First, practitioners {\it do} want to express functional constraints, as partial identification is notoriously uninformative in many cases. Shape constraints may be fruitful even in purely predictive tasks \citep[e.g][]{gupta:2020}, as further background knowledge to reduce degrees of freedom that cannot be fully decided by data alone. Many domains have no clear reason to allow for dose-response curves that have many inflection points, and in fact the community has been criticized by creating benchmarks with artificial outcome functions for the sake of illustrating overly complex average treatment effect estimators, which hardly deliver any advantages in several empirical domains \citep[e.g.,][]{gentzel:2021}. The argument that restricting function spaces may lead to invalid bounds is vacuous and misses the point of causal modeling. A model is what we remove from the space of all possibilities, that is, the constraints we adopt.\footnote{To use a quote usually apocryphally attributed to Michelangelo: ``It is the sculptor’s power, so often alluded to, of finding the perfect form and features of a goddess, in the shapeless block of marble; and his ability to {\bf chip off all extraneous matter}, and let the divine excellence stand forth for itself.'' \emph{The Methodist Quarterly Review} (1858), emphasis added.} The missing edges in a causal graph are just that: constraints. Invalid bounds are a consequence of an overconstrained feasible region, and the function space is part of the game as much as any decisions about missing edges. The promotion of a type of constraint (independence constraints) as having a protected place in causal modeling should not be taken seriously by methodologists or practitioners.

Second, like missing edges, function spaces can be partially tested. Testability is predicated on assumptions that themselves may be untestable: in the case of purely missing edges, faithfulness \citep{sgs:2000} and its variants; in the case of function spaces and the independence constraints that go along with them, we can test whether our mathematical program has a feasible solution or not. Passing the mathematical program is not a guarantee of correctness -- which is not surprising, given that e.g. hypothesis testing and many frameworks of scientific falsifiability are about deciding not to reject a postulated model, not about confirming it. As another example, passing Fisher's sharp null hypothesis by itself does not prove that individual treatment effects are exactly, or even close to, zero (or whether counterfactuals have a physical meaning at all!), and so on. Nevertheless, there are important applications of falsifying a causal model with partial identifiability \cite[e.g][]{wolfe2019inflation,robins:2015}, which are useful so long as it is understood that ``testability'' is always an asymmetric concept.

Among the practical uses of constrained function spaces, for instance, is the possibility of expressing knowledge as smooth deviances from responses that do not control for unmeasured confounding: e.g., parameterizing $Y = f_{(w, u_Y)}(x)$ as $Y = \mathbb E[Y~|~w, x] + r_{(w, u_Y)}(x)$, the latter function $r_{(w, u_Y)}(x)$ taking particularly smooth shapes. 

One line of criticism that deserves more serious consideration was raised by \cite{dawid:03} in the context of classical categorical models for instrumental variables: postulating constraints on structural equations/response functions is equivalent to making counterfactual claims which are never directly testable, even with perfect randomized controlled trials (RCTs). This motivated Dawid to create an alternative to partial identification in discrete IV settings without postulating the existence of response functions. A discussion of interventional (response-function-free) and counterfactual (adopting response functions) models is provided by \cite{swanson:2018}. In our case, expressing models in terms of latent variables and conditional distributions falls back to the issue we discussed in the main text, by which it is unclear how to interpret  an infinite dimensional latent variable space and its relation to the conditional distributions, and all the computational complications that follow. Our take is that the response functions are idealizations that do not need to correspond to counterfactuals if all that is required is how well we can model interventional distributions -- themselves testable implications in the sense of comparisons against RCTS. Without RCTs, the formulation has a degree of falsifiability in terms of the feasible region of the mathematical program, as discussed above.

\section{A general implementation of the SCP}
\label{app:general-scp}

We show here how we can instantiate the SCP for general graphs. Recall that we observe data from a causal model~$S^{\star}$, with density function $p_{S^\star}$ over continuous variables, including a continuous treatment~$X \in \bR^p$ and outcome~$Y \in \bR$ among the observed variables. Let $o(\mathcal S^*)$ be a causal estimand of interest. In our case, this is the functional $\E_{S^{\star}}[Y \given do(X=x^{\star})]$. Our goal is to minimize/maximize $o_{x^{\star}}(S^{\star})$ over all \emph{admissible} causal models $S$ among a (uncountable) model class $\cS$.

In \cref{sec:method}, we describe a smart parametrization of the causal model where we only need to estimate the first two moments of a conditional distribution. This was possible by taking advantage of the structure of the problem and baking some of the constraints directly into the construction of the generative model. The advantage of this is that the optimization is more stable and reliable, as we have seen in \cref{sec:experiments}. Alternatively, we can use any off the shelf generative model to define our model class, $\cS$ as we describe in the following subsection.

\subsection{Parametrizing the causal model}
We write $V = g_{V, \boldsymbol{\theta}_V}(\mathrm{Pa}_V)$, where $\boldsymbol{\theta}_V$ are the parameters of our response functions. This means that we need to optimize over a distribution over $\boldsymbol{\theta}$. 

We model each endogenous variable $V$ as a normalizing flow conditioned on $\mathrm{Pa}_V$. We denote the flow transformation corresponding to $V$ as $g_V$ and the base distribution as $s_v$. We start with the root endogenous variables. Since they do not have endogenous parents, they are simply flow transformations of the noise corresponding to the c-component. We can learn these as fixed flows that we then sample from to get the generated variable. Samples generated through this architecture for a variable $V$ are denoted by $V_g$. Alternatively, for endogenous variables $V$ without any endogenous parents (meaning $\mathrm{Pa}_V$ is an empty set) we can also simply use the observations of $V$ as $V_g$ to represent a trivial mapping. We continue downstream in this way, with each variable being a flow conditioned on the generated parents, so $V_g = g_{V, \boldsymbol{\theta}_V}(s_v | {(\mathrm{Pa}_V)}_g)$.\footnote{This is strictly speaking an abuse of notation. We write it like this to make the point that $s_v$ is the noise being transformed, and it is done conditional to having the values of the generated parents ${(\mathrm{Pa}_V)}_g$ given as input as well.} Repeating this architecture for each endogenous $V$ will give us a generative model for our endogenous variables. So in the end, the parameters of our optimization procedure will be $\eta = \{\theta_V\}_{\{V \in \mathcal{V}\}}$ (consistent with the notation from \cref{sec:method}). This generative model corresponds to $p_S$.

Note that this construction is similar to the one described in \citet{hu2021}. The difference is that we still manage to avoid a bi-level optimization by simply using any differentiable difference between distribution measure as $\dist$ instead of using a discriminator as \citet{hu2021} do. We also note that flows are not at all essential to this general implementation, other generative models can easily be used as well. This is simply one instantiation of how the generic SCP might be implemented using off the shelf methods.

Next, we show how we enforce the structural constraints. 
\subsection{Encoding structural constraints}
We note that $\boldsymbol{\theta}_V$ inherits the independence relationships between $U_V$. We use the c-component factorization of \citet{tian2002general}, since we know that any pair of exogenous variables are independent if they belong to different c-components \citep{tian2002general}. We share the same noise term among flows corresponding to variables within the same c-component. This ensures that the independence assumptions are encoded.

A small technicality here is that flows require the base distribution to be the same dimension as the output dimension, but variables in the same c-component will in general not be of the same dimension. Taking the example of the IV setting, $X \in \mathbb{R}^p$ and $Y \in \mathbb{R}$ are in the same c-component, meaning that they have a shared base distribution.To get around this, we use a base distribution of dimension $p$ and use the first dimension of this distribution as the base for $Y$. In practice, this means that we draw samples from the $p$-dimensional base distribution and use the first dimension of these $p$-dimensional samples as the base samples for $Y$. Note that this still respects the independence assumptions of the model. We can do a similar process in general even with more variables being in the same c-component, using a base distribution with the dimension equal to the dimension of the maximum dimensional variable among those in the same c-component.

\xhdr{Implementation} 
In our implementation, we use conditional affine coupling flows \citep{dinh2016density} as our conditional flows, with an implementation in PyTorch provided by the python library Pyro \citep{bingham2019pyro}. As the $\dist$ we use MMD \citet{muandet:2017}, which will enforce the data constraints. The objective value $o(\mathcal S)$ can be calculated from the generative model. For example, by using the output of the flows we are using as part of our generative model, except that the flow corresponding to the intervened variable is replaced by a constant value. Finally, this leads to a constrained optimization as defined in \cref{eq:optimization} with a single constraint. This is finally solved using the augmented lagrangian method, as described in \cref{app:optimization}. We use a slack of $0.02$ to account for the scale of the MMD, which is different from the slack used in the clustered SCP implementations (\cref{app:implementation}). For the MMD, we use $1000$ samples from the generative model and use $1000$ randomly chosen points from the observed distribution to compare the distance.

\section{Extensions}
\label{app:extensions}

We introduced our problem formulation as a very generic definition of model proximity in terms of some $\dist(p_\eta, \hat p)$, and in particular the formulation
\begin{equation*}
\dist(p_\eta, \hat p) \equiv ||f^{p_\eta} - f^{\hat p}||_{\infty},
\end{equation*}
\noindent where $f^{\hat p}$ is a finite-dimensional vector of functionals of the (estimated) observable distribution, and $f^{p_\eta}$ is a finite-dimensional vector of functionals of the distribution as implied by an unidentifiable structural causal model. \cref{eq:constraintentry} is a particular implementation of this idea. Due to the challenge of evaluating this norm, we suggest Monte Carlo approximation for the optimization procedure, based on \citet{wang2016stochastic}.

We have also presented results for alternative distance metrics, including special cases such as metrics that can be associated with a single Lagrange multiplier, such as $||f^{p_\eta} - f^{\hat p}||_2$. We note that also combinations of a limited number of metrics are possible such as
\begin{equation*}
\max \{||f^{p_\eta}_1 - f^{\hat p}_1||_1, ||f^{p_\eta}_1 - f^{\hat p}_2||_1, \dots, ||f^{p_\eta}_K - f^{\hat p}_K||_1\},
\end{equation*}
where each $k = 1, 2, \dots, K$ describes a cluster of functionals that are easy to compare on the same scale.

A possible extension is the provision of a calculus for a minimal parameterization of arbitrary structural causal models for a given causal query of interest. Looking back at our instrumental variable scenario, we could have created a single set of latent variables $N_{XY}$ that would be common causes of $X$ and $Y$ and model the structural equations for both $X$ and $Y$ at the same time, instead of fixing $p(X \given Z)$ a priori and deterministically extracting latent variables $N$ from $X$ and $Z$. In general, we could add independent latent variables for each clique in the bidirected graph component of the causal graph. This, however, is very wasteful. There is no need to create a causal model for the $(Z, X)$ marginal -- which not only would assume that $Z$ is an unconfounded cause of $X$, which is unnecessary, but would also waste computation and stability trying to match the observable marginal of $(Z, X)$ to the corresponding causal-model-implied marginal, which is also completely unnecessary. Given that $Z$ and $X$ can be high-dimensional while $Y$ is a scalar, this is clearly a bad idea. Therefore, the structure-blind strategy of creating latent variables for each bidirected clique is convenient but not ideal, and a smarter automated way of generating minimal causal parameterizations for arbitrary graphs and causal queries is needed.

\section{Dataset description}
\label{app:data}

\subsection{Semi-synthetic data: Simulating Mendelian randomization}
\label{app:mandelian-randomization}

\begin{wrapfigure}{l}{0.43\textwidth}
    \vspace{-5pt}
    \centering
    \textbf{Semi-synthetic yeast data} \\
        \includegraphics[width=1\linewidth]{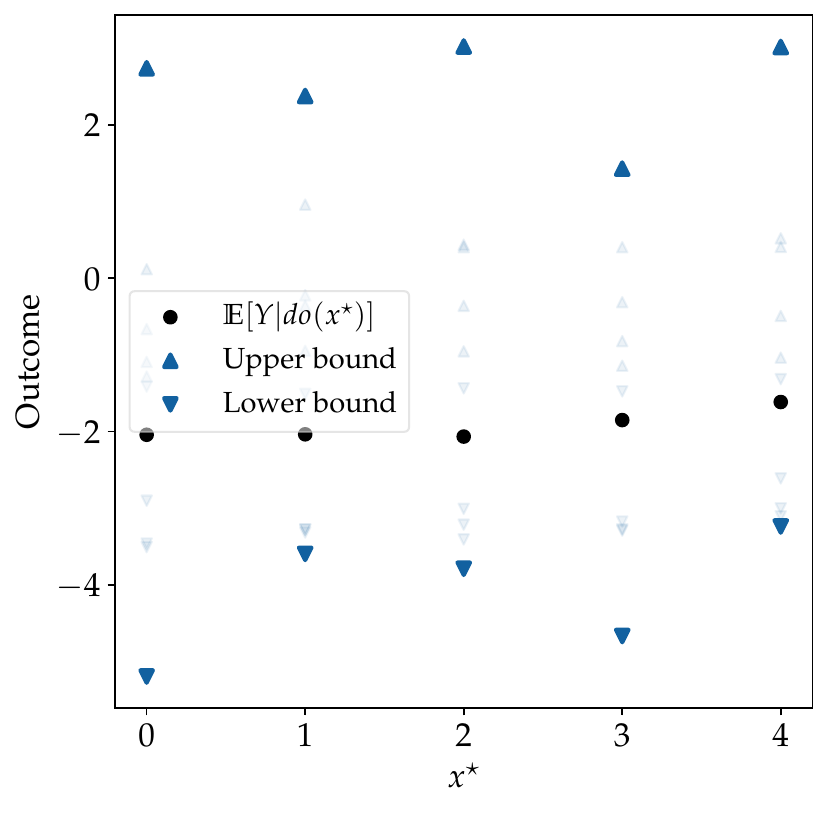}
    \caption{\textbf{Simulating mendelian randomization}. We get valid bounds even with very few samples.}
    \vspace{-21pt}
    \label{fig:app-iv-yeast}
\end{wrapfigure}
We use a semi-simulated setup inspired by Mendelian randomization (MR). MR studies exploit the natural genetic variation induced by meiosis to quantify the causal effect of phenotypes on health outcomes, even when the two are subject to latent confounding. In this setting, the genotype serves as an instrumental variable. Though we can never be certain that the untestable IV assumptions hold, MR methods are widely used in genetic epidemiology \citep{didelez2007}.

For our semi-synthetic setting, we use data from an expression quantitative trait loci (eQTL) study of \emph{Saccharomyces cerevisiae}, where the goal is to identify genetic sources of variation in mRNA expression. The data include 112 $\text{F}_1$ segregants, a cross of parental strains BY4716 and the wild isolate RM11-1a \citep{Brem2005}. We choose $5$ eQTLs on chromosome $14$, which is known to regulate the mitochondrial ribosomal network in \textit{S. cerevisiae}, as our instruments ($Z$). The $15$ genes representing this ribosomal network are then the treatments ($X$). Finally, a synthetic outcome is generated as a function of $X$ and a latent confounder $C$. The treatment is therefore $15$-dimensional, and the instrument is $5$-dimensional. Treatments are continuous, while instruments are binary (indicating presence or absence of an allele). Strong dependencies are evident among both sets of variables. 

This data poses several challenges to our method, since (a) there are very few samples; (b) there are strong dependencies between sets of features; and (c) the instruments are not continuous. Despite these challenges, our method manages to find valid bounds as we see in \cref{fig:app-iv-yeast}.

\xhdr{Generation of the outcome} We use samples from a $5$-dimensional multivariate standard normal $c = (c_1, c_2 \ldots c_5)$ as confounders. We choose a vector $\beta = (\beta_1, \beta_2 \ldots \beta_{15})$ of length $15$, with each $\beta_i$ chosen uniformly at random from $[0, 1]$ (but fixed once chosen). Note that $x = (x_1, x_2\ldots x_{15})$.  Finally, we set $Y = \sum_{i = 1}^{5} \beta_i * c_i * x_i + \sum_{i = 6}^{15} \beta_i * x_i + n_Y$. Here $n_Y$ is a standard Gaussian noise. For the xstar values, we choose a random $x$ from the treatments, say $x_0$. Then to get $x^*_i$, we add random noise to three indices of $x_0$, namely indices $3*i + 1$, $3*i + 2$ and $3*i + 3$. So we get five xstars since the indices go up to $15$ (the dimension of $X$) where each $x^*_i$ has three indices shifted from a point in the dataset. This is done to ensure that the intervention values are not completely unrealistic given the observed data. In summary, the outcome is a linear function of the treatment, with non-additive confounding.

\xhdr{Implementation details} For the implementation, we use the clustered SCP formulation and the $l_\infty$ norm. We choose to have $112$ data points as constraints, corresponding to each point in the sample. This gives us a total of $224$ constraints, since we have two constraints for each point (the first and second moment). We use the linear response function family, meaning in this case that we use the polynomial basis with $15$ dimensions. Otherwise, we use the same parameters as described in \cref{app:implementation}.

\subsection{Glossary of Synthetic Datasets}
We describe here the synthetic datasets we use in our experiments.

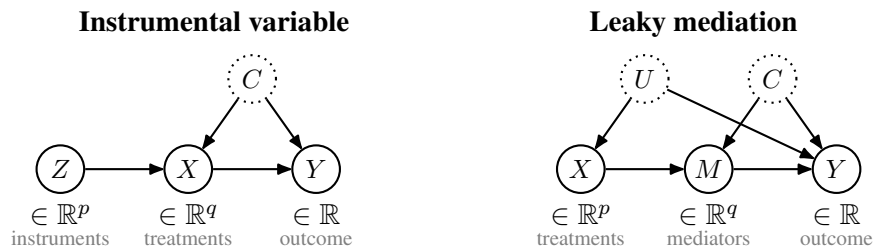
\begin{figure}[ht]
    \centering
  \phantom{}\hspace{-0.2cm} \textbf{Instrumental variable} \hspace{3cm} \textbf{Leaky mediation} \\[2mm]
    \begin{tikzpicture}
    \node[obs,label={[align=center]below:{$\in \bR^p$ \\[-2mm]\scriptsize {\color{gray}instruments}}}] (Z) at (0, 0) {$Z$};
    \node[obs,label={[align=center]below:{$\in \bR^q$ \\[-2mm]\scriptsize {\color{gray}treatments}}}] (X) at (1.7, 0) {$X$};
    \node[obs,label={[align=center]below:{$\in \bR$ \\[-2mm]\scriptsize {\color{gray}outcome}}}] (Y) at (3.4, 0) {$Y$};
    \node[latent] (C) at (2.55, 1.2) {$C$};
    \edge[black]{Z}{X};
    \edge[black]{X}{Y};
    \edge[black]{C}{X};
    \edge[black]{C}{Y};
  \end{tikzpicture}
  \hspace{2cm}
  \begin{tikzpicture}
    \node[obs,label={[align=center]below:{$\in \bR^p$ \\[-2mm]\scriptsize {\color{gray}treatments}}}] (X) at (0, 0) {$X$};
    \node[obs,label={[align=center]below:{$\in \bR^q$ \\[-2mm]\scriptsize {\color{gray}mediators}}}] (M) at (1.7, 0) {$M$};
    \node[obs,label={[align=center]below:{$\in \bR$ \\[-2mm]\scriptsize {\color{gray}outcome}}}] (Y) at (3.4, 0) {$Y$};
    \node[latent] (U) at (0.85, 1.2) {$U$};
    \node[latent] (C) at (2.55, 1.2) {$C$};
    \edge[black]{X}{M};
    \edge[black]{M}{Y};
    \edge[black]{U}{X};
    \edge[black]{U}{Y};
    \edge[black]{C}{M};
    \edge[black]{C}{Y};
  \end{tikzpicture}
  \caption{These are the structural equations we work with. In the description of the datasets, the dimensions of each variable are denoted by a subscript. For instance, if $X$ is $2$-dimensional, we write it as $X = (X_1, X_2)$.}
  \label{fig:app-scm}
\end{figure}

\begin{figure}[ht]
    \centering
    \tikz[baseline]{
            \node[anchor=base, draw, line width=0.1pt] (t1)
            {IV};
        } -
        \tikz[baseline]{
            \node[anchor=base, draw, line width=0.1pt] (t2)
            {lin};
        } -
        \tikz[baseline]{
            \node[anchor=base, draw, line width=0.1pt] (t3)
            {2d}
        } -
        \tikz[baseline]{
            \node[anchor=base, draw, line width=0.1pt] (t4)
            {strong}
        } -
        \tikz[baseline]{
            \node[fill=gray!20, anchor=base] (t5)
            {add (optional)}
        }
    \caption{Naming logic for the datasets. The \textit{first} segment mentions whether it is the Instrumental Variable (IV) or Leaky Mediator (LM) setting. The \textit{second} segment says whether $y$ is linear or quadratic in $x$. The \textit{third} segment tells the dimension of the treatment. The \textit{fourth} segment denotes the strength of the confounding, strong or weak. The \textit{fifth}, and last segment is optional, and says whether the confounding is additive, which in the IV setting makes the effect identifiable.}
    \label{fig:app-naming-logic}
\end{figure}

We use various polynomial datasets where the causal effect is a polynomial function of a single or multidimensional treatment. We provide the construction of each of these here. \cref{fig:app-scm} shows the structural graphs. In both graphs, every node except $Y$ (outcome) could be multi-dimensional. If a node, say $C$, is multi-dimensional, we index the dimension with a subscript. That is, we write $C = (C_1, C_2)$. \cref{fig:app-naming-logic} visually describes the naming logic behind the datasets to make the exposition clearer.

\subsubsection{Scalar treatment}
We now describe settings where the treatment $X$ is a scalar, which we present as a sanity check. The noises, confounder, and instrument follow
\begin{equation*}
 e_X, e_Y, c, z \sim \cN(0, 1)
\end{equation*}

\begin{itemize}[leftmargin=*]
\item \texttt{IV-lin-1d-weak-add} ($f_y$ linear in $x$, weak additive confounding)
\begin{align*}
    f_X(z, c, e_X) &= 3 z + 0.5 c + e_X \\
    f_Y(x, c, e_Y) &= x - 6c + e_Y
\end{align*}

\item \texttt{IV-quad-1d-strong} ($f_y$ quadratic in $x$, strong non-additive confounding)
\begin{align*}
    f_X(z, c, e_X) &= 0.5 z + 3 c + e_X \\
    f_Y(x, c, e_Y) &= 0.3 x^2 - 1.5 x c + e_Y
\end{align*}

\item \texttt{IV-quad-1d-weak} ($f_y$ quadratic in $x$, weak non-additive confounding)
\begin{align*}
    f_X(z, c, e_X) &= 3 z + 0.5 c + e_X \\
    f_Y(x, c, e_Y) &= 0.3 x^2 - 1.5 x c + e_Y
\end{align*}
\end{itemize}

\subsubsection{IV model}

We now describe datasets satisfying the IV assumptions.

\xhdr{2D treatment}
The noises, confounder, and instruments follow
\begin{align*}
c, z, e_X &\sim \cN^2(0, 1)\\
e_Y &\sim \cN(0, 1)
\end{align*}

\begin{itemize}[leftmargin=*]
\item  \texttt{IV-lin-2d-strong} ($f_y$ linear in $x$, strong non-additive confounding)
\begin{align*}
    f_X(z, c, e_X) &= 0.5 z + 2 c + e_X \\
    f_Y(x, c, e_Y) &= x_1 +  x_2 - 3 (x_1 + x_2)(c_1 + c_2) + e_Y
\end{align*}

\item \texttt{IV-lin-2d-weak} ($f_y$ linear in $x$, weak non-additive confounding)
\begin{align*}
    f_X(z, c, e_X) &= 2 z + c + e_X \\
    f_Y(x, c, e_Y) &= 5x_1 +  6x_2 - x_1(c_1 + c_2) + e_Y
\end{align*}

\item \texttt{IV-quad-2d-strong-add} ($f_y$ quadratic in $x$, strong additive confounding)
\begin{align*}
    f_X(z, c, e_X) & = z + 2 c + e_X \\
    f_Y(x, c, e_Y) & =2 x_1^2 + 2 x_2^2 - (c_1 + c_2) + e_Y
\end{align*}

\item \texttt{IV-quad-2d-weak} ($f_y$ quadratic in $x$, weak non-additive confounding)
\begin{align*}{2}
    f_X(z, c, e_X) &= 2 z + c + e_X \\
    f_Y(x, c, e_Y) &= 5x_1^2 +  6x_2^2 - (x_1+x_2)(c_1 + c_2) + e_Y
\end{align*}
\end{itemize}

\xhdr{3D treatment}
The noises, confounders, and instruments follow
\begin{align*}
c, z, e_X &\sim \cN^3(0, 1)\\
e_Y &\sim \cN(0, 1)
\end{align*}
\begin{itemize}[leftmargin=*]
    \item \texttt{IV-quad-3d-weak} ($f_y$ quadratic in $x$, weak non-additive confounding)
    \begin{align*}
        f_X(z, c, e_X) &= 2 z + c + e_X \\
        f_Y(x, c, e_Y) &= 2 x_1^2 + 2 x_2^2 + 2 x_3 \\
                       &\phantom{=} - 0.3 (x_2+x_3)(c_1 + c_2 + c_3) + e_Y
    \end{align*}
 \end{itemize}
  
\subsubsection{Leaky Mediator}
We now describe datasets following the leaky mediator model with noises, and confounder following
\begin{align*}
e_X, e_M, c, u &\sim \cN^2(0, 1)\\
e_Y &\sim \cN(0, 1)
\end{align*}

\begin{itemize}[leftmargin=*]
    \item \texttt{LM-lin1-2d} ($f_y$ linear in $x$, strong confounding)
    \begin{align*}
        f_X(u, e_X) &= u + e_X \\
        f_M(x, c, e_M) &= x + 3c - e_M \\
        f_Y(m, c, u, e_Y) &= 2m_1 + m_2\\
                         &\phantom{=} - (m_1 + m_2)(c_1 + c_2 + u_1 + u_2) + e_Y
    \end{align*}

    \item \texttt{LM-lin2-2d} ($f_y$ linear in $x$, weak confounding)
    \begin{align*}
        f_X(u, e_X) &= u + e_X \\
        f_M(x, c, e_M) &= 3x + c - e_M \\
        f_Y(m, c, u, e_Y) &= 2m_1 + m_2 \\
                         &\phantom{=} - 0.3(m_1 + m_2)(c_1 + c_2 + u_1 + u_2) + e_Y
\end{align*}
\end{itemize}

\section{Implementation Details}
\label{app:implementation}

We use $n=10\,000$ data points for all our simulations of synthetic datasets. All implementation is in Python, using PyTorch~\citep{pytorch}. Code for the experiments is provided with the supplementary material. An overview of the algorithm, specialized for the IV case, is shown in Algorithm \ref{algo:method}.

\begin{algorithm*}
\caption{Computing upper or lower bounds on $\E[Y\given do(X=x^{\star})]$ in the IV setting.}\label{algo:method}
\begin{algorithmic}[1]
{\small
  \Require
  dataset $\cD = \{(z_i, x_i, y_i)\}_{i=1}^n$;
  constraint functions $\{\phi_l: \cY \to \bR \}_{l=1}^L$;
  basis functions $\{\psi_k: \cX \to \cY\}_{k=1}^K$;
  norm $\|\cdot\|$ for $\dist$;
  batchsize for Monte Carlo $B$;
  number of support points $M$;
  tolerance $\epsilon > 0$
  \Statex \commentbox{\textbf{Setup:} One-time computations shared for different $x^{\star}$ values}
  \State Fit (invertible) conditional normalizing flow $X = h_Z(N)$ from data $\cD$ for $N \sim \cN(0, \B{1}_p)$%
  \State Fit MLPs $\hat{\phi}_1: x_i, z_i \to y_i$ and $\hat{\phi}_2: x_i, z_i \to y_i^2$ by minimizing the squared loss from data $\cD$%
  \State subsample~$M$ indices from $[n]$ (uniform, no replacement) \Comment{``support points'', w.l.o.g.\ use $[M]$}
  \Statex \commentbox{\textbf{Optimization:} performed separately for lower and upper bound for each $x^{\star}$}
  \State minimize \Call{Objective}{$\eta$} subject to \Call{Constraint}{$\eta$}${}\le \epsilon$ (see \cref{app:optimization})
  \Statex
  \Function{Objective}{$\eta$}
    \State $o_{x^{\star}}(\eta) \gets \psi(x^{\star})^\top \frac{1}{B} \sum_{j = 1}^B \mu_{\eta_0}(n_j)$\quad with $n_j \sim \cN(0, \B{1}_p)$ \Comment{differentiable w.r.t.\ $\eta$}
    \State \Return $\pm o_{x^{\star}}(\eta)$ \Comment{objective, $\pm$ for lower/upper bound}
  \EndFunction
  \Function{Constraint}{$\eta$}
    \State $n_j \gets h_{z_j}^{-1}(x_j)$ for $j \in [M]$ \Comment{invert $X \given Z$ model to infer ``noises''}
    \State $\rhs_{1,j}(\eta) \gets \psi(x_j)^\top \mu_{\eta_0}(n_j)$ for $j \in [M]$ \Comment{moments implied by model}
    \State $\rhs_{2,j}(\eta) \gets \psi(x_j)^\top \Bigl(\Sigma_{\eta_1}(n_j) + \mu_{\eta_0}(n_j) \mu_{\eta_0}(n_j)^\top \Bigr)  \psi(x_j)$ for $j \in [M]$%
    \State $\nu_{l,j} \gets \hat{\phi}_l(x_j, z_j) - A_{l,j}(\eta)$ for $l \in \{1, 2\}, j \in [M]$ \Comment{constraint matrix}
    \State \Return $\|\nu\|$ 
  \EndFunction
}
\end{algorithmic}
\end{algorithm*}

\subsection{Satisfying \cdata{}} 
In \cref{subsec:data} we explained how we match the observed data distribution by matching scalar statistics for $L$ dictionary functions  $\{\phi_l: \cY \to \bR \}_{l=1}^L$. In practice, we take $L=2$ and match the first and second moments of $Y \given\{X, Z\}$, which is to say that we set $\phi_1(Y) = Y$ and $\phi_2(Y) = Y^2$. We learn $\phi_1$ and $\phi_2$ by regressing MLPs on the observed data $\{x_i, z_i, y_i\}_{i\in [n]}$ and, $\{x_i, z_i, y_i^2\}_{i\in [n]}$ respectively, with  $\{x_i, z_i\}_{i\in [n]}$ as the input and  $\{y_i\}_{i\in [n]}$ and  $\{y_i^2\}_{i\in [n]}$ as the target values. The outputs of these MLPs are then approximations of $\E[Y\given X, Z]$ and $\E[Y^2\given X, Z]$, respectively. A significant advantage of this approach is that we can evaluate the constraints in closed form under our construction (see \cref{eq:rhs-form}). We use MLPs with $3$ hidden layers of sizes $(64, 32, 16)$ (first to last) and train it with a batch size of $512$ for $200$ epochs, with a learning rate of $0.01$. We use ReLU activations. Identical settings were used for the leaky mediator, with the only difference being that we match $Y \given\{M, X\}$ instead of $Y \given\{X, Z\}$.

\subsection{Satisfying \cstruct{}}
We considered two distinct ways of modeling $X\given Z$. The same idea is analogous when modeling $M~|~X$

\xhdr{Gaussian}. We write the mean and variance of $X$ as a function of $Z$. That is to say, $X = h_Z(N) := A(Z) N + b(Z)$. We parameterize $A(Z)$ as $A = L^\top L + \Omega$. A Cholesky factor $L: \cZ \to \bR^{p \times p}$ ensures symmetry, and a diagonal matrix $\Omega$ adds small constants to the diagonal to ensure symmetry and positive definiteness.
We then learn $L^\top L + \Omega$ from observed data $\cD$ by maximizing the log-likelihood of $\hat{p}(X \given Z)$. In practice, the parameters of $A(Z)$ as described above, and the parameters of $b(Z)$ (which are just the entries of the vector $b(Z)$) are the output of an MLP which takes in $X$. We learn the weights of this neural net once up front from observed data $\cD$ by maximizing the log-likelihood of $\hat{p}(X \given Z)$. In practice, this MLP has $3$ hidden layers of sizes $(64, 32, 16)$ (first to last) and is trained with a batch size of $512$ for $200$ epochs, with a learning rate of $0.01$. We use ReLU activations.

\xhdr{Conditional normalizing flow}. We use an invertible (conditional) normalizing flow to model the distribution of $X\given Z$. Flows are a natural candidate for modeling distributions, and in this case follow both the properties we want from the modeling of $X\given Z$. (i) Given $X, Z$, we can invert the transformation to get $N$. (ii) We can sample from $X \given Z=z$. We use the Python library Pyro \citep{bingham2019pyro}. 

All the experiments shown in this work use conditional normalizing flows due to better performance and stable optimization. However, the \emph{Gaussian modeling} is described to point out one more way in which prior assumptions on functional dependencies could be incorporated in our generic framework. Once again, the construction is identical for the leaky mediator setting. The only difference is that there we model $M \given X$ instead of $X \given Z$.

\subsection{Parameterizing \texorpdfstring{$\cS$}{S}}
We consider $\theta$ to be $K$-dimensional and use the \textit{neural basis function} in all experiments unless specified otherwise. \cref{eq:thetadistr} describes how we parameterize our model. Here $\theta \given N$ is parameterized by two MLPs with parameters  with weights denoted by $\eta_0$ and $\eta_1$. The weights of this MLP constitute our main optimization parameters $\eta$ and we denote the outputs by $\mu_{\eta_0}$ and $\Sigma_{\eta_1}$. $\mu_{\eta_0}$ has 2 hidden layers with size $(16, 16)$ and $\Sigma_{\eta_0}$ has 2 hidden layers with size $(32, 32)$. To evaluate the objective function, we perform Monte Carlo estimation $\E_{N}[\mu_{\eta_0}(N)]$ with 500 samples for $N$ as in \cref{eq:constrainedfinal}.

The difference in the leaky mediator setting is that the parameterization of $\theta$ has 2 noise inputs instead of 1. The remaining implementation choices remain unchanged.

\subsection{Response function choice}
\label{app:response}
Our choice of response functions is characterized by the choice of $K$ basis functions, as we saw in \cref{eq:f_theta}. We primarily use the \textit{neural basis functions} described in \cref{subsec:implementation}. Note that a linear combination of basis functions allows us to be arbitrarily expressive with our choice of family of response functions. 
In particular, we consider the following options:
\begin{enumerate}[leftmargin=*,topsep=0pt,itemsep=0pt]
  \item \emph{Neural basis functions:} The basis function $\psi_k$ is the activation of the $k$th neuron in the last hidden layer of an MLP which has been trained on the observed data $\{x_i, y_i\}_{i\in [n]}$ to learn $y$ given $x$.
  
  In practice, we train a 3-hidden layer MLP with 64 neurons in the first two layers, rectified linear units (ReLU) as activation functions and an MSE (mean squared error) loss for $100$ epochs and a batch size of $512$ using Adam with a learning rate of $0.01$. The size $K$ of the last hidden layer is equal to the number of basis functions we wish to have in our family of response functions.
  
  An implicit assumption we have here is that the MLP can find a good approximation of $y$ given $x$. This assumption is well-supported by theoretical and empirical results showing the approximation power of neural networks~\citep{shen2021neural, daubechies2021nonlinear}, making it the perfect candidate for multi-dimensional treatments.
  
  \item \emph{Adaptive basis functions:} The basis function $psi_k$ is $\tanh(w_{0k} + w_{1k}x)$. The key point here is that the weights here are not fixed, but rather also included as parameters to optimize over while finding the min / max in the constrained optimization. This gives a semantic separation to $f_{\theta}$ since $\theta$ depends on $X$ and $Z$ only via the distribution of $U$, while $\psi$ is a black box function of $X$. Due to the adaptive nature of the weights $\{w_{0i}, w_{1i}\}_{i\in[K]}$, we would expect this choice of basis to give wider bounds than the neural basis functions. This is illustrated through experiments in \cref{fig:app-iv-lin-2d-adaptive}.
  
  \item \emph{Polynomials:} For a multivariate input $x=(x_1, x_2\ldots x_d)$, a response function can be considered to be a multivariate polynomial in $x$. However, in this case, we would have $2^d$ different basis functions of degree up to $d$. This leads to the question of what basis functions to use here to have enough expressive power, but not blow up the number of basis functions and hence the dimension of the optimization parameter $\theta$ (see \cref{eq:f_theta}). Due to this blowup in the choice of basis functions with the dimension of the treatment in the case of polynomials, mainly use the neural basis functions.
  
  In the cases when we do use the polynomial basis, we restrict ourselves to having up to quadratic terms in the basis function. For a $d$-dimensional treatment, this amounts to having a $d(d + 3)/2 + 1$ dimensional $\theta$.
  
  \item \emph{Gaussian process basis functions (GP):} We do not use this family of response functions, but mention them as a possible option. There is some previous work which considers the GP basis to define a family of functions~\citep{kilbertus2020class}. In this approach, a Gaussian process is fit to $K$ different sub-samples $\{(x_i, y_i)\}_{i \in N'}$ with $N' \le N$.
  A single function is then sampled from each Gaussian process as the basis functions $\psi_k$ for $k \in [K]$. Here, `sampling a function' means to get the evaluation of the function at several points in the treatment space and then interpolate. While this is a reasonable approach when the treatment is scalar, for a multidimensional treatment, this requires the interpolating a multivariate function. The complexity and computational cost of such an interpolation increases with the increase in dimensionality, while the reliability of the interpolated function decreases at the same time. Also, higher dimension require exponentially higher number of point evaluations to get good interpolations. Due to these reasons, the GP basis approach is not found suitable for higher dimensions. However, it can be a viable option for scalar treatments.
\end{enumerate}

\begin{figure*}[!ht]
    \centering
    \textbf{Identifiable settings}  \\
    \begin{subfigure}{0.48\textwidth}
        \centering
        Scalar treatment \smallskip
        \includegraphics[width=1\linewidth]{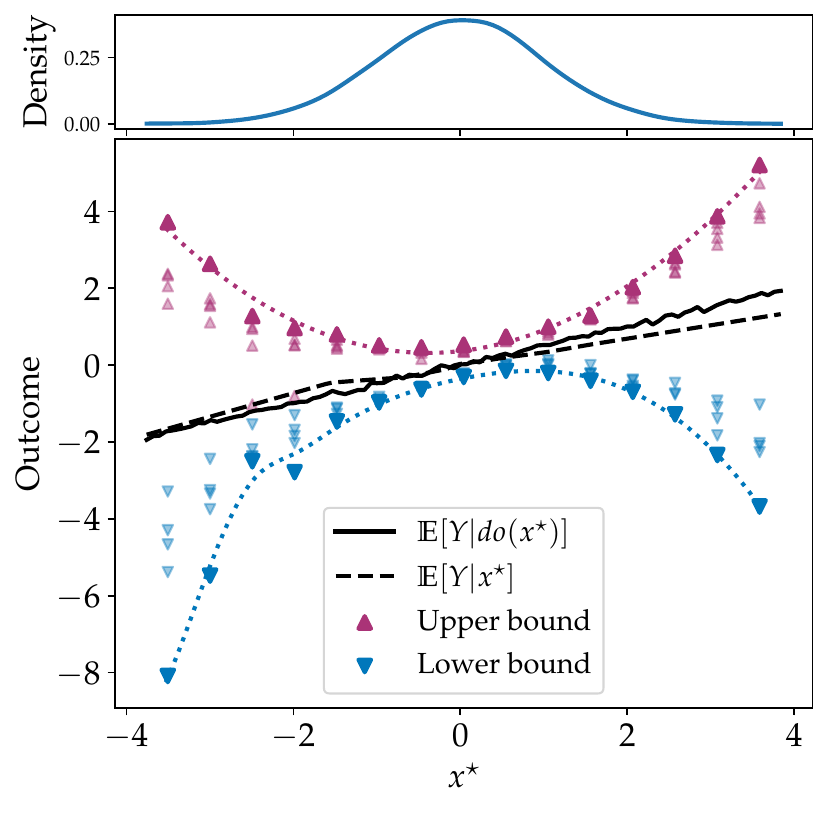}
        \caption{\textbf{IV-lin-1d-weak-add}. An identifiable scalar treatment with weak confounding. We get tight bounds despite using the neural basis functions.}
    \end{subfigure}\hfill
    \begin{subfigure}{0.48\textwidth}
        \centering 
        Multi-dimensional treatment \smallskip
        \includegraphics[width=1\textwidth]{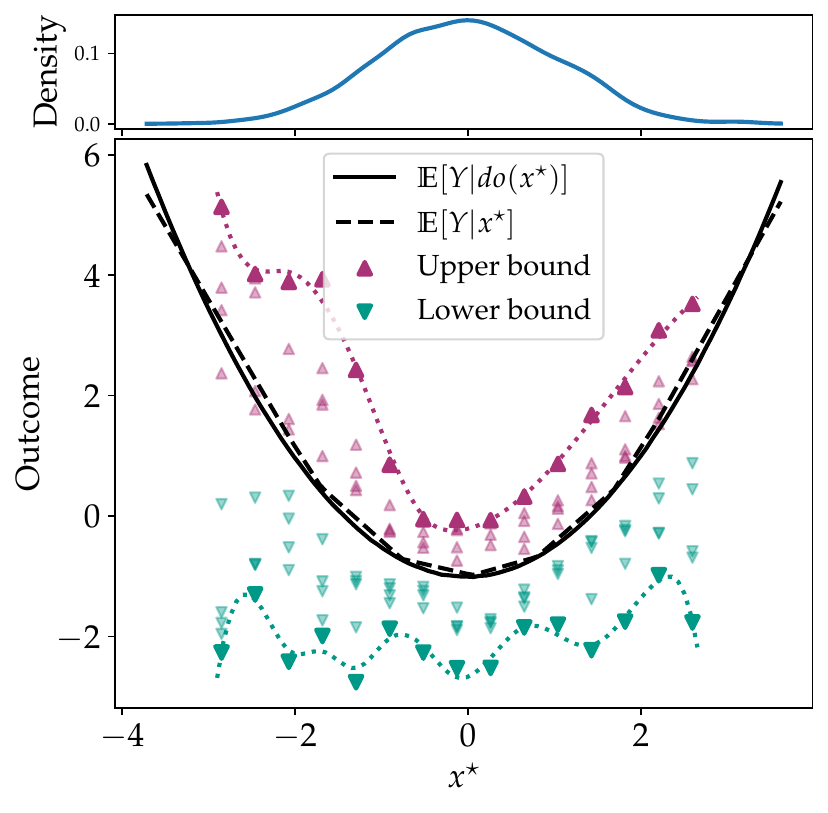}
        \caption{\textbf{IV-quad-2d-strong-add}. Multi-dimension treatment using the neural basis functions.}
    \end{subfigure}
    \label{fig:app-multi-identifiable}
    \caption{We see that our method is able to find tight bounds in identifiable settings in the data-dense regions.}
\end{figure*}

\begin{figure*}[!ht]
    \centering
    \textbf{Strong instrument VS Weak instrument} (Scalar treatment) \\
    \begin{subfigure}{0.48\textwidth}
        \centering
        \includegraphics[width=1\linewidth]{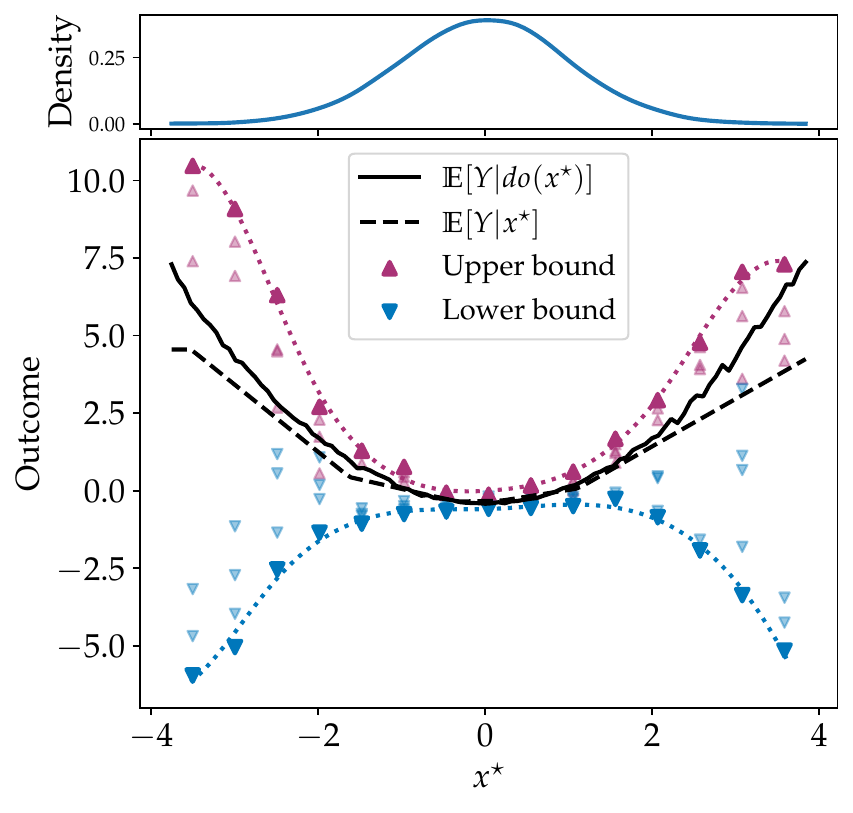}
        \caption{\textbf{iv-quad-1d-weak}. This is a partially identifiable scalar treatment setting. The bounds are particularly tight in the data-dense regions due to the strong instrument.}
    \end{subfigure}\hfill
    \begin{subfigure}{0.48\textwidth}
        \centering
        \includegraphics[width=1\textwidth]{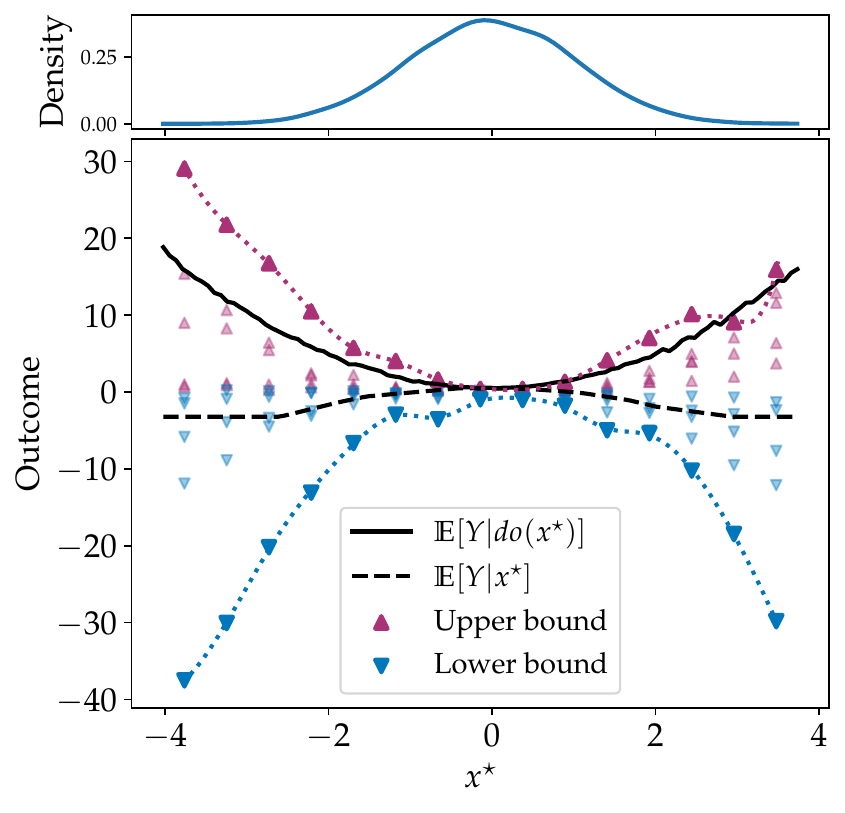}
        \caption{\textbf{iv-quad-1d-strong}. This is a partially identifiable scalar treatment setting. The bounds are looser because the instrument is weak.}
    \end{subfigure}
    \caption{As expected, the bounds are tighter in the case of strong instrument.}
    \label{fig:app-comp-conf-scalar}
\end{figure*}

\subsection{Solving the optimization} \label{app:optimization}
The various MLPs described in the previous sections eventually lead to the formulation of a constrained optimization problem that we have seen in \cref{eq:constrainedfinal}.

We use the augmented Lagrangian method for inequality constraints to solve the constrained optimization problem in \cref{eq:constrainedfinal}. The formulation is taken from Section 17.3 in \citep{nocedal2006numerical}.

We have seen in \cref{subsec:optimization} that we have a total of $D \cdot L$ constraints. We can think of $\hat{\phi}_l(x_i,y_i)$ as target values, estimated once up front from observed data. We denote $\hat{\phi}_l(x_i,y_i) = \lhs_{l,i}$.
The right-hand side $\rhs_{l,i}$ is a function of the optimization parameter $\eta$, as seen in \cref{subsec:data}. For ease of notation, we ``flatten'' the indices $n$ and $l$ into a single index $l \in [D\cdot L]$. We set the constraint slack to be the same value for each constraint, so $\epsilon_L = \epsilon$.

Then our set of constraints is
\begin{equation*}
  c_l(\eta) :=  \epsilon - |\lhs_l - \rhs_l(\eta)| \geq 0
\end{equation*}

With this, the Lagrangian we aim to minimize with respect to $\eta$ can be formulated as:
\begin{equation}\label{eq:subproblem}
  \cL(\eta, \lambda, \tau) := \pm o_{x^{\star}}(\eta) + \sum_{l=1}^{D \cdot L} \xi(c_l(\eta), \lambda_l, \tau)
\end{equation}
with
\begin{equation*}
  \xi(c_l(\eta), \lambda_l, \tau) :=
  \begin{cases}
    - \lambda_l c_l(\eta) + \frac{\tau c_l(\eta)^2}{2}
      & \text{if } \tau c_l(\eta) \le \lambda_l,  \\
    - \frac{\lambda_l^2}{2 \tau}
      & \text{otherwise},
  \end{cases}
\end{equation*}
where $-$/$+$ is used for the upper/lower bound. $\tau$ is increased throughout the optimization procedure and is seen as a temperature parameter.

Given an approximate minimum $\eta$ of this subproblem, we then update $\lambda$ and $\tau$ according to $\lambda_l \gets \max\{0, \lambda_l - \tau c_l(\eta)\}$ and $\tau \gets \alpha \cdot \tau$ for all $l \in [D \cdot L]$ and a fixed $\alpha > 1$.
The overall strategy is to iterate between minimizing \cref{eq:subproblem} and updating $\lambda_l$ and $\tau$.
We find empirical justification in our experiments, where the approach reliably converges on a range of different datasets.
We summarize our proposed procedure in \cref{algo:method}.

We have already seen how the basis function is chosen, how we can write the $\rhs$ of the constraints in closed form, how the $\lhs$ of the constraints is fixed upfront using $\phi_1$ and $\phi_2$, and how the objective is estimated. It now remains, to solve this optimization problem, which we solve using the augmented lagrangian method as described in S. In practice we fix the initial value of $\tau$ to be $\tau_{init}=10$, and the multiplier $\alpha$ to be $4$, meaning that $\tau$ is updated by multiplying it by $4$ at each step. The max value of $\tau$ is fixed at $\tau_{max}=5000$. We use $30$ optimization steps to find the approximate optimal $x_k$ at the $k^{th}$ round of the optimization, and perform $150$ round of optimization (number of times $\lambda$ is updated) for each value of $x^{\star}$ and each bound (upper and lower). The optimization was performed using the Adam optimizer with a learning rate of $0.001$. All of this again is through the use of auto-differentiation in PyTorch \citet{pytorch}, implemented in python.

\begin{figure*}[!ht]
    \centering
    \textbf{GAN comparison} \\
    \begin{subfigure}{0.48\textwidth}
        \centering
        \includegraphics[width=1\linewidth]{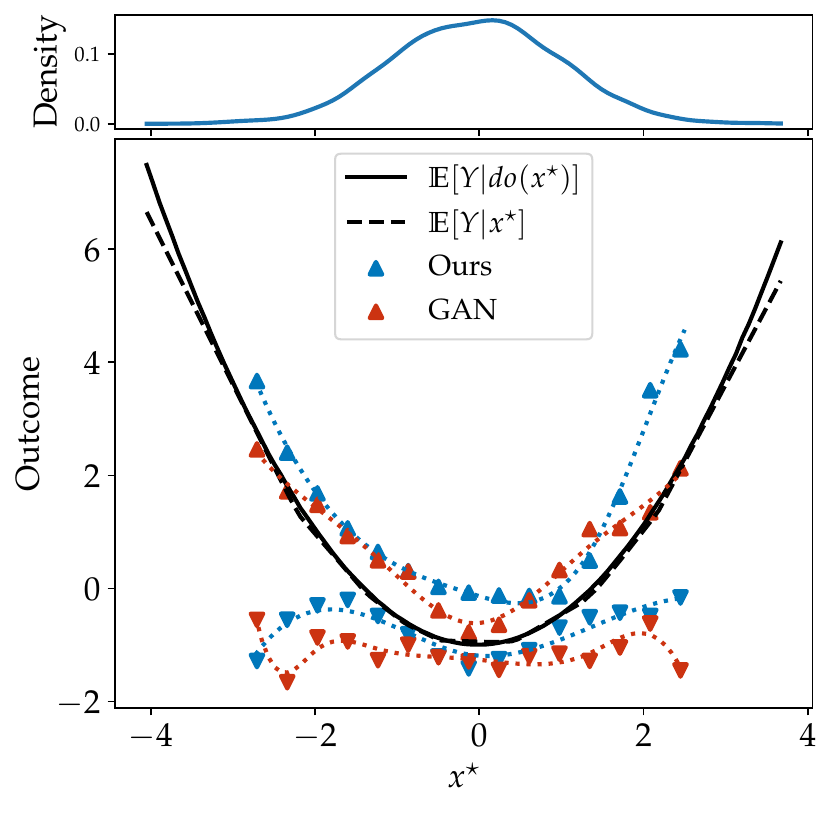}
        \caption{\textbf{IV-quad-2d-weak}. }
    \end{subfigure}
    \begin{subfigure}{0.48\textwidth}
        \centering
        \includegraphics[width=1\textwidth]{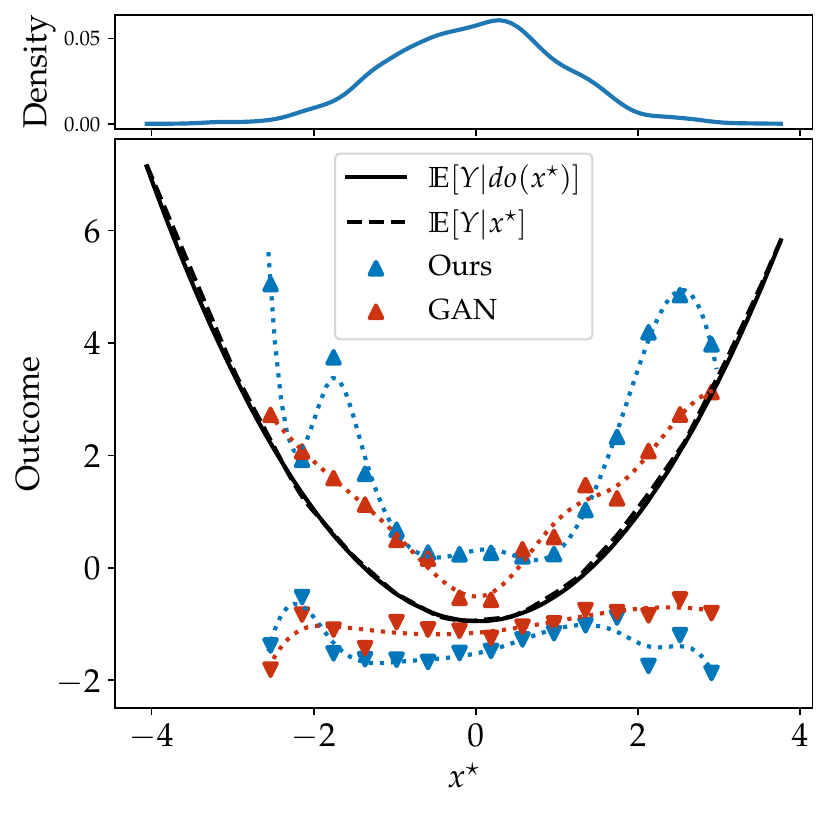}
        \caption{\textbf{IV-quad-3d-weak}.}
        \label{app:fig-iv-quad-3d-weak}
    \end{subfigure}
    \caption{We do note that in this instance, the bounds given by our method are not as smooth in the data poor regions. However, our bounds are always valid, while the GAN framework gives some invalid bounds in both cases above.}
    \label{fig:app-gan-compare}
\end{figure*}

\subsection{Hyperparameter search}
We did not perform an automated hyperparameter search for $\tau_{init}$, $\tau_{max}$ and $\tau_{factor} = \alpha$ since we observed that the values $\tau_{init}=10$, $\tau_{max}=100000$ and $\tau_{factor} = 5$ worked reasonably well in all our settings. We arrived at this value through trial and error. An automated hyperparameter search can also be expected to improve the bounds.

\subsection{Computational resources}
We used a computing cluster provided by the university, for ease of parallelization of experiments. We used 1 CPU, 8000 MB of RAM for all the $x^{\star}$ bounds for a single random seed for the method. To clarify, the use of the cluster was only for the purpose of running the optimization algorithm parallelly for various different random seeds. A single run of the algorithm (for getting bounds on multiple values of $x^{\star}$) runs comfortably on a local machine with 16 GB of RAM. No GPUs were used for the experiments.

\section{Further Experiments} \label{app:more-experiments}
Here we show the performance of our method in a variety of settings.

\begin{figure*}[!t]
    \centering
    \textbf{Norm comparison} \hspace{45mm} \textbf{Adaptive basis} \\
    \begin{subfigure}{0.48\textwidth}
        \centering
        \includegraphics[width=1\linewidth]{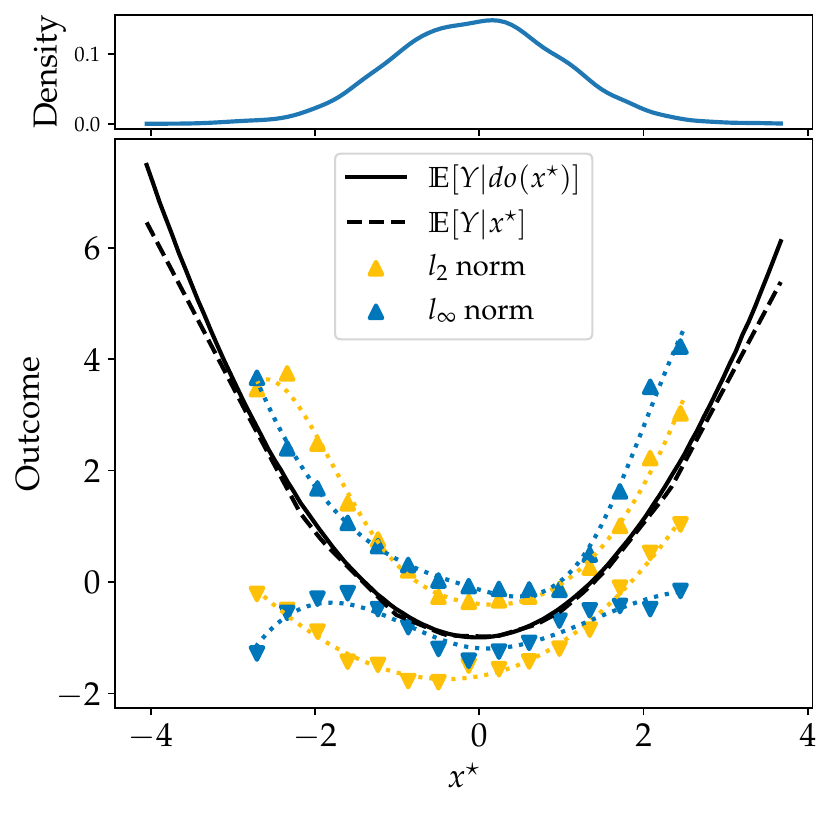}
        \caption{\textbf{IV-quad-weak}. Partially identifiable multi-dimension treatment. The $l_2$ norm includes all the constraints without sampling.}
        \label{fig:app-comp-norm}
    \end{subfigure}\hfill
    \begin{subfigure}{0.48\textwidth}
        \centering
        \includegraphics[width=1\textwidth]{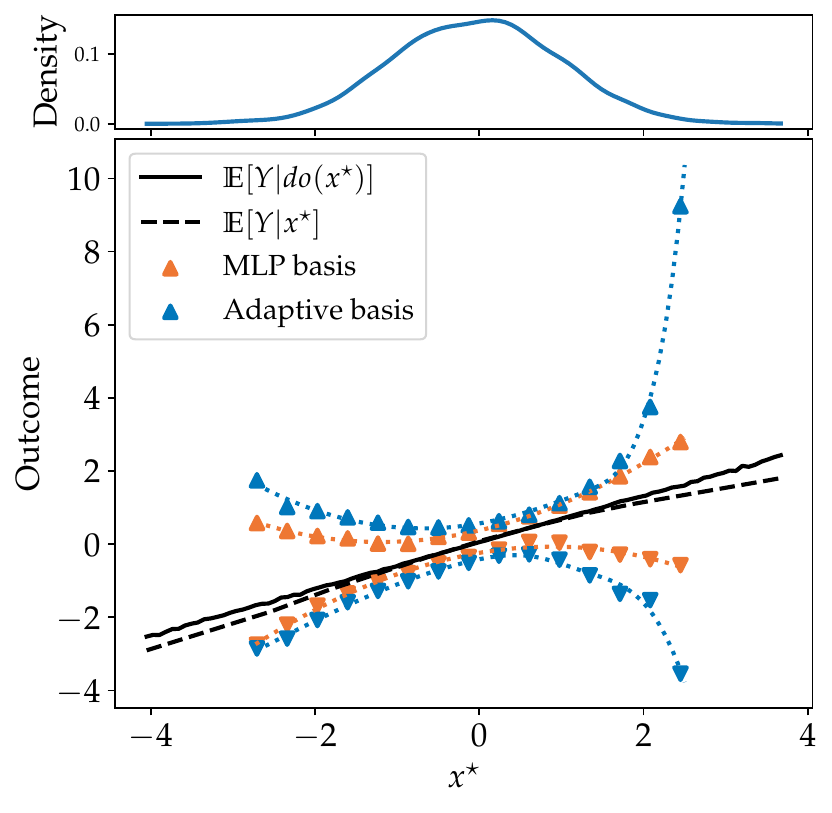}
        \caption{\textbf{IV-lin-2d-weak}. We see that the the adaptive basis gives looser bounds than the MLP basis, as we would expect.}
        \label{fig:app-iv-lin-2d-adaptive}
    \end{subfigure}
    \label{fig:app-comp-norm-adaptive}
    \caption{Some more comparisons}
\end{figure*}

\subsection{Identifiable settings}
\cref{fig:app-multi-identifiable} shows the performance in the case of identifiable settings. We see that our method is able to find tight bounds in identifiable settings in the data dense regions. However, it is natural that the more expressive the response function, the wider the bounds. This can potentially be used as a test of identifiability as described in \cref{sec:limitations}. Such a test can be seen as a rather continuous measure of identifiability, giving an indication of being somewhere between identifiable and partially identifiable based on the tightness of bounds, rather than being able to distinct strictly between identifiable and non-identifiable settings. 

\subsection{Strong instrument VS Weak instrument}
\cref{fig:app-comp-conf-scalar} compares the bounds in the cases of having a strong or weak instrument, or equivalently having weak or strong confounding relative to the instrument. As expected, the bounds are looser when the confounding is stronger.

\subsection{Comparing constraint norms}
In \cref{subsec:data} we have formulated our constraints in terms of norms. All the experiments so far have been using the $\|\cdot\|_{\infty, \infty}$ norm. Here, we show experiments using the $\|\cdot\|_{2,2}$ norm as a  constraint. This optimization procedure remains the same, but is potentially easier to solve since there is only $1$ constraint now, but on the other hand this constraint is looser than matching the moment of each data point (or a chosen subsample). The results can be seen in \cref{fig:app-comp-norm}.

We point out that our formulation does not restrict us to norm based constraints. In fact, given the generative model we have defined in \cref{sec:method}, we could also use any measure of distance between distributions as a constraint. However, this would again lead to solving a bi-level optimization, making the optimization harder. 

\begin{figure}[!ht]
    \centering
    \textbf{GAN comparison} \\
    \begin{subfigure}{0.48\textwidth}
            \centering
            \includegraphics[width=1\linewidth]{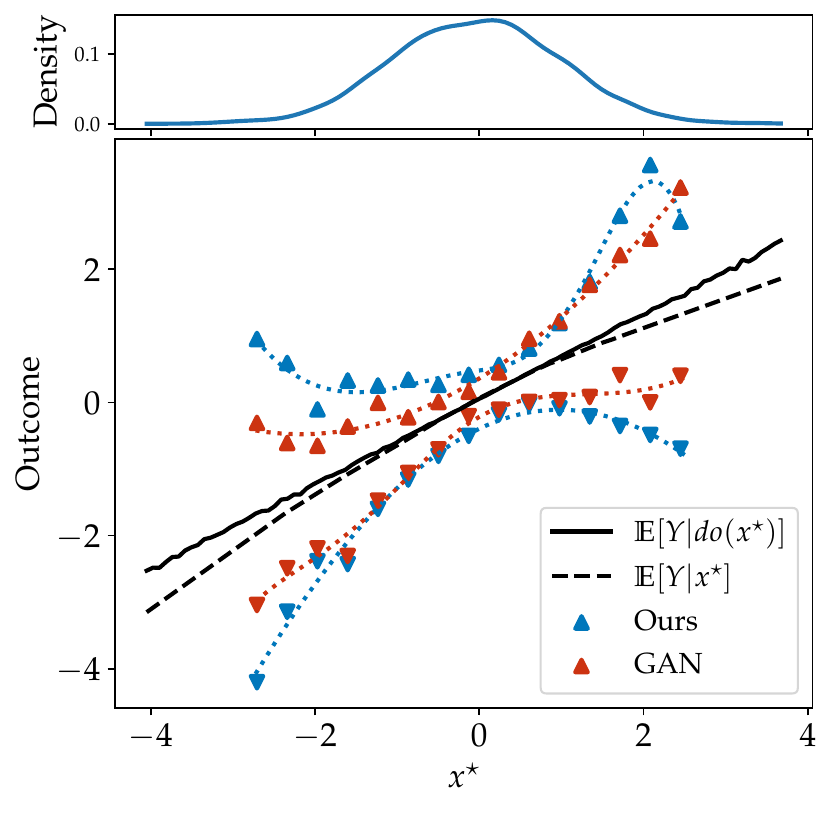}
            \caption{\textbf{IV-lin-2d-weak}. }
            \label{fig:app-gan-compare-2}
    \end{subfigure}
    \begin{subfigure}{0.48\textwidth}
            \centering
            \includegraphics[width=1\linewidth]{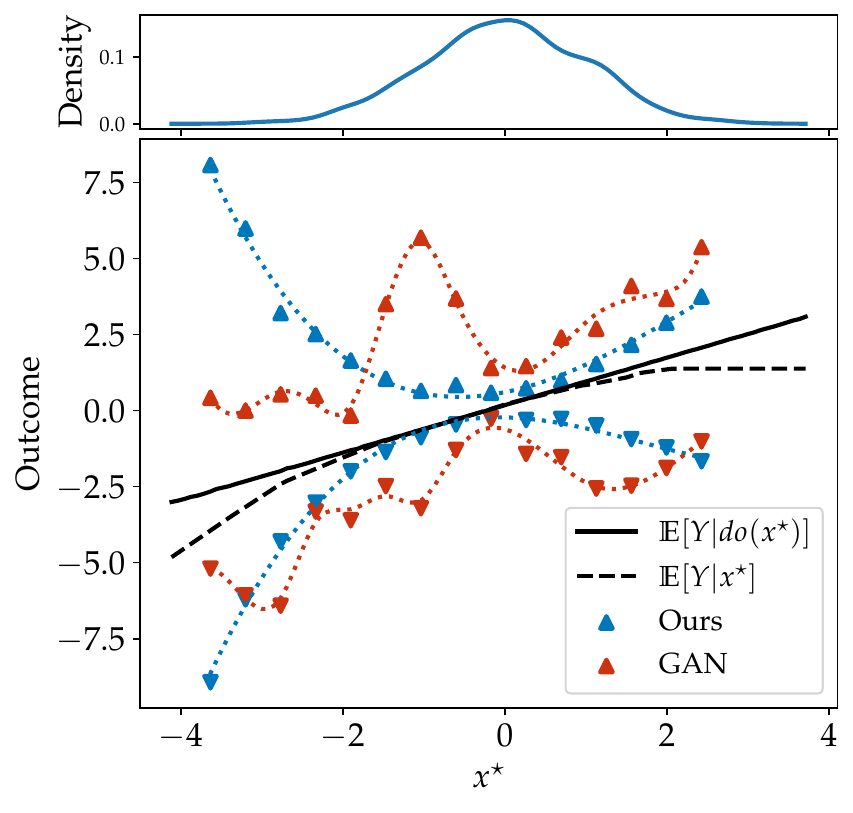}
            \caption{\textbf{LM-lin-2d-weak}. }
            \label{fig:app-gan-compare-fd}
    \end{subfigure}
    \caption{Comparison with GAN}
\end{figure}

\subsection{More comparisons to GAN bounds}
We show some further comparisons of our method to the GAN bounds here. We see a similar trend where our bounds are generally smoother. The plots can be found in \cref{fig:app-gan-compare}, \cref{fig:app-gan-compare-2} and \cref{fig:app-gan-compare-fd}.

\section{Limitations of \texorpdfstring{\citet{kilbertus2020class}}{Kilbertus et al. (2020)}}
\label{app:limiations}

\citet{kilbertus2020class} limit the number of constraints by estimating $\E[\phi_l(Y) \given z_i]$ only for a small pre-determined set of grid points, $z_i$ around which they bin the observed datapoints for empirical estimates.
Thereby, they only constrain $p(Y \given Z)$ instead of $p(Y \given X, Z)$ and fundamentally limit themselves to low-dimensional settings, since otherwise the number of grid points grows exponentially with the dimension of $\cZ$.
In particular, their proposed interpolation of empirical cumulative distribution functions to fix $p(x \given z)$ in their copula model that parameterizes the distribution over $\theta$ virtually only allows one-dimensional instruments and treatments.
Note that the binning procedure for $Z$ also becomes problematic in the small data regime, where sufficiently many points are needed in each bin to keep the variance in the empirical mean and variance estimates low.
As a consequence, the number of gridpoints in $z$ must be carefully tuned depending on the dataset size.

Instead, our stochastic subsampling approach works with a fixed number of constraints for each update step, independent of the size of the dataset and the dimensionality of $Z$ and $X$.
In addition, we encode the structural assumptions into a graphical model for $X, Z, \theta$ that still allows for flexible conditional density estimation techniques such as invertible flows for individual components.
The copula model had to rely on interpolated empirical cumulative density function estimated from potentially few datapoints within each bin.
The Gaussian copula is also less flexible than our proposal.
Finally, \citet{kilbertus2020class} used Monte Carlo estimates both for the objective and for each individual constraint, often using different sample sizes for these estimates.
Instead, we exploit the form of response functions as linear combinations of basis functions and compute the constraints in closed form, removing the additional variance from the stochastic optimization procedure.
This leads to fewer tunable parameters and more robust convergence of the optimization.

\section{Ethical and social implications} \label{app:ethics}
The ethical implications of causal inference are numerous, especially when machine learning models are used to make high-stakes decisions in areas like finance, criminal justice, and healthcare. Several authors in recent years have developed causal notions of algorithmic fairness to acknowledge the structural links between protected attributes and socially significant outcomes \cite{kusner2017, kilbertus2017avoiding, wu2019b, kilbertus2019sensitivity}. More reliable methods for bounding causal effects could help quantify the counterfactual fairness of individual decisions when identifiability is impossible. By the same token, these bounds may misrepresent treatment effects when untestable assumptions are not satisfied. In the worst case, an agent may design a DAG in an adversarial manner to ensure some desired result. This can be mitigated by public disclosure of all structural assumptions, so that regulators and data subjects may critically evaluate decision procedures.

\end{document}